\documentclass[arxiv]{iosart2x}

\usepackage{mathptmx}
\usepackage{hyperref}
\usepackage{amsmath}
\usepackage{graphicx}
\usepackage{xspace}
\usepackage{color}
\usepackage{amssymb}

\newcommand{\change}[1]{#1}

\newcommand{\ie}{\textit{i}.\textit{e}.,\xspace}
\newcommand{\eg}{\textit{e}.\textit{g}.,\xspace}
\newcommand{\cf}{\textit{cf}.\xspace}

\renewcommand\vec{\mathbf}
\newcommand{\mat}[1]{\mathbf{#1}}
\newcommand{\func}[1]{\text{#1}}
\newcommand{\norm}[1]{\left\lVert#1\right\rVert}
\newcommand{\inner}[1]{\left\langle#1\right\rangle}

\usepackage{multirow}

\usepackage{graphicx}
\usepackage{xspace}
\usepackage{color}
\usepackage{enumerate}
\usepackage{paralist}
\usepackage{hyperref}
\usepackage[draft]{minted} 
\usepackage{caption}
\usepackage{listings}

\usepackage{array}
\newcolumntype{C}[1]{>{\centering}m{#1}}

\usepackage{hhline}
\usepackage{footnote}
\makesavenoteenv{tabular}
\makesavenoteenv{table}

\usepackage{float}
\usepackage{listings}

\newfloat{code}{htbp}{lop}
\floatname{code}{Listing}

\usepackage{subcaption}

\usepackage{amsmath,amssymb}
\usepackage{tikz-cd}
\newcounter{rowcntr}[table]
\renewcommand{\therowcntr}{\texttt{$t_{\arabic{rowcntr}}$}}

\newcolumntype{N}{>{\refstepcounter{rowcntr}\therowcntr}c}

\AtBeginEnvironment{tabular}{\setcounter{rowcntr}{0}}

\pubyear{2022}
\volume{13}
\issue{3}
\firstpage{1}
\lastpage{40}

\begin{document}
\makeatletter
\let\put@numberlines@box\relax
\makeatother
\begin{frontmatter}              

\title{Prediction of Adverse Biological Effects of Chemicals Using Knowledge Graph Embeddings}
\runtitle{Prediction of Adverse Biological Effects of Chemicals Using Knowledge Graph Embeddings}


\author[A,B,F]{\fnms{Erik} \snm{B. Myklebust}%
\thanks{Corresponding Author: Erik B. Myklebust, NORSAR, Gunnar Randers Vei 15, 2007 Kjeller, Norway. E-mail: \href{mailto:erik@norsar.no}{erik@norsar.no}}}
\author[C,B]{\fnms{Ernesto} \snm{Jim\'{e}nez-Ruiz}\thanks{Corresponding Author: Ernesto Jiménez-Ruiz, City, University of London, College Building, Northampton Square,
London EC1V 0HB. E-mail: \href{mailto:ernesto.jimenez-ruiz@city.ac.uk}{ernesto.jimenez-ruiz@city.ac.uk}}},
\author[D]{\fnms{Jiaoyan} \snm{Chen}},
\author[A,G]{\fnms{Raoul} \snm{Wolf}}
and
\author[A,E]{\fnms{Knut Erik} \snm{Tollefsen}}

\runauthor{E. B. Myklebust et. al.}
\address[A]{Norwegian Institute for Water Research, Oslo, Norway}
\address[B]{SIRIUS, University of Oslo, Oslo, Norway}
\address[C]{City, University of London, London, United Kingdom}
\address[D]{University of Oxford, Oxford, United Kingdom}
\address[E]{Norwegian University of Life Sciences, \AA s, Norway}
\address[F]{NORSAR, Kjeller, Norway}
\address[G]{Norwegian Geotechnical Institute, Oslo, Norway}


\begin{abstract}
We have created a knowledge graph based on major data sources used in ecotoxicological risk assessment. We have applied this knowledge graph to an important task in risk assessment, namely chemical effect prediction. \change{We have evaluated nine knowledge graph embedding models from a selection of geometric, decomposition, and convolutional models on this prediction task. We show that using knowledge graph embeddings can increase the accuracy of effect prediction with neural networks. 
} 
Furthermore, we have implemented a fine-tuning architecture which adapts the knowledge graph embeddings to the effect prediction task and leads to a better performance. \change{Finally, we evaluate certain characteristics of the knowledge graph embedding models to shed light on the individual model performance.}

\end{abstract}

\end{frontmatter}

\section{Introduction}


Ecotoxicology is a multidisciplinary field that studies the potentially adverse toxicological effects of chemicals on organisms, starting at molecular level to individuals, sub-populations, communities and ecosystems. One major societal contribution of ecotoxicology 
is ecological risk assessments, which compare environmental concentrations of chemicals with existing laboratory effect data to evaluate the ecosystem health status. While laboratory experiments are thus crucial, they are both labour intensive and result in a high number of animal testing. Therefore, the development of modelling techniques for extrapolating from existing laboratory effect data is a major effort in the field of ecotoxicology. 

A very important challenge in ecotoxicology risk assessment is the interoperability of the 
disparate data sources, formats and vocabularies. The use of Semantic Web technologies and (RDF-based) knowledge graphs \cite{ARNAOUT201866} can address this challenge and 
facilitate the orchestration of these datasets. Hence, 
extrapolation or prediction models can benefit from an integrated view of the data and the background knowledge provided by a knowledge graph.
%
The use of knowledge graphs also enables the use of the
available infrastructure to perform automated reasoning, explore the data via semantic queries, and compute semantic embeddings for machine learning prediction.

In this work we have created
the Toxicological Effect and Risk Assessment Knowledge Graph (TERA) and implemented a prediction model over this knowledge graph to extrapolate adverse biological effects of chemicals on organisms. Here, we limit ourselves to \change{binary effect prediction of mortality (shortened to effect prediction), \ie where there is a chance that a chemical can affect a species in a lethal way.
}
\change{
The work and evaluation conducted in this paper is driven by the following research question: \emph{does the use of contextual information in the form of knowledge graph embeddings brings added value in the  prediction of adverse biological effects?}
}

Our contributions can be summarized as follows:
\vspace{-0.2cm}
\begin{enumerate}[\it (i)]
    \item 
    TERA aims at consolidating the relevant information to the ecological risk assessment domain. TERA integrates several disparate datasets and enables a unified (semantic) access. The formats of these data sources 
    vary from tabular, to RDF files and SPARQL 
    endpoints over public linked data. 
    We have exploited external resources (\eg Wikidata \cite{wikidata2014}) and ontology alignment methods (\eg LogMap~\cite{logma_ecai2012}) to discover equivalences between the data sources.
    
    \item We have designed and implemented a model tailored to binary lethal chemical effect prediction. This model relies on TERA and builds upon existing knowledge graph embedding models. Moreover, it supplies the knowledge graph embedding models with additional information. This is used to tailor the embeddings to this specific task. 
    
    \item We have evaluated nine knowledge graph embedding (KGE) models, together with a naive baseline on the binary chemical effect prediction task. 
    This evaluation includes four data sampling strategies which highlight the different settings of chemical effect prediction (\ie the test data contains unseen chemical-organism pairs where: \textit{(a)} the chemical and the organism may be known (but not in previously seen pairs), \textit{(b)} the chemical is unknown, \textit{(c)} the organism is unknown, and  \textit{(d)} both the chemical and the organism are unknown). 
\end{enumerate}
\change{
These contributions are openly shared. A snapshot of the TERA knowledge graph is available on Zenodo~\cite{myklebust_erik_bryhn_2019_4244313} (\url{https://doi.org/10.5281/zenodo.3559865})
and the source scripts for creating TERA are available on GitHub (\url{https://github.com/NIVA-Knowledge-Graph/TERA}). Finally, the scripts to reproduce the conducted evaluation in this paper are also available on GitHub (\url{https://github.com/NIVA-Knowledge-Graph/KGs_and_Effect_Prediction_2020}).}

This paper extends our preliminary work presented in the In-Use Track of the 18th International Semantic Web Conference \cite{Myklebust2019KnowledgeGE}. 
We have 
\begin{inparaenum}[\it (i)]
    \item extended TERA with new sources (Encyclopedia of Life (EOL), MeSH, and a larger part of ChEMBL) and provided
    detailed steps about its creation;
    \item created a more robust prediction model with \change{nine (up from three)} embedding algorithms supported and a task-specific embedding fine-tuning strategy; and
    \item conducted a more comprehensive evaluation with \change{all combinations of KGE models} and sampling strategies \change{totalling 648 data points (324 for each prediction model).}
\end{inparaenum}

The rest of the paper is organized as follows.
\change{Section \ref{sec:preliminaries} introduces essential concepts to the subsequent sections.}
Section \ref{sec:niva} introduces the use case where the knowledge graph and prediction models are applied. Section \ref{sec:related} introduces related work. The creation of the knowledge graph is described in Section \ref{sec:tera}. Section \ref{sec:prediction} introduces the prediction models, while Section \ref{sec:results} presents the evaluation of these models. Section~\ref{sec:discussion} elaborates on the contributions and discusses future directions of research. Finally, Appendix \ref{sec:appendix_kge} gives an overview of the knowledge graph embedding models used in this work.

\section{Preliminaries}
\label{sec:preliminaries}
\change{
In this section we introduce important background concepts that will be used throughout the paper. Table~\ref{tab:symbols} contain the most important symbols.

\begin{table}[]
    \centering
    \begin{tabular}{|c|l|}
        \hline 
         Symbol & Definition \\\hline 
         RDF & Resource Description Framework\\
         OWL & Web Ontology Language\\
         SPARQL & SPARQL Protocol and RDF Query Language\\
         KG & Knowledge graph\\
         KGE & Knowledge graph embedding \\
         $t$ & A triple \\
         $sb$ & The subject of a triple\\
         $ob$ & The object of a triple\\
         $p, r$ & The predicate/relation of a triple \\
         
         $e$ & A KG entity \\
         $\mathcal{T}$ & The set of KG triples \\ 
         $\mathcal{E}$ & The set of KG entities \\ 
         $\mathcal{R}$ & The set of KG relations \\
         $\mathcal{L}$ & The set of literal values \\
         $\vec{e}$ & The vector representation of an entity or relation\\
         k & The dimension of a vector\\
         $SF$ & The scoring function of a KGE model \\
         $PT$ & Pre-trained KGE-based model \\ 
         $FT$ & Fine-tuning KGE-based model \\ 
         $s$ & A species \\
         $c$ & A chemical \\
         $S$ & Refers to species \\
         $C$ & Refers to chemicals \\
         $\kappa$ & Chemical concentration \\
         
         \hline
    \end{tabular}
    \medskip
    \caption{Key symbols and acronyms used throughout the paper.}
    \label{tab:symbols}
\end{table}

\subsection{Ecotoxicological terminology}

\textit{Taxonomy} in this work refers to a species classification hierarchy. Any node in a taxonomy is called a \textit{taxon}. \textit{Species} is a taxon which is also a leaf node in the taxonomy. An \textit{Organism} denotes an individual living organism which is an instance of a species.
\textit{Chemicals or compounds} are unique isotopes of substances consisting of two or more atoms.
\textit{Effect}, used in this work as short form for chemical effect, refers to the response of an organism (or population) to a chemical at a specific concentration. 
\textit{Endpoint}\footnote{Not to be confused with SPARQL endpoint.} denotes a measured effect on the test population at a certain time; \eg lethal concentration to 50\% of test population (LC50) measured at 48 hours. Note that, an experiment can have several endpoints, \eg LC50 at 48 hours and LC100 at 96 hours (lethal concentration for all test organisms). See Table \ref{tab:endpoints} for the most common endpoints.

\subsection{Ontology-enhanced knowledge graphs}
In this work we consider the most broadly accepted notion of knowledge graph within the Semantic Web: an ontology enhanced RDF-based knowledge graph~(KG)~\cite{DBLP:journals/corr/abs-2003-02320}.
This kind of knowledge graph enables the use of the available Semantic Web infrastructure, including SPARQL engines and OWL reasoners.\footnote{RDF, RDFS, OWL and SPARQL are standards defined by the W3C: \url{https://www.w3.org/standards/semanticweb/}}
Thus, in our setting, KGs are composed by RDF triples in the form of $\left\langle sb, p, ob \right\rangle \in \mathcal{E}\times\mathcal{R}\times\mathcal{E} \cup \mathcal{L}$,\footnote{$\mathcal{E}$ is the set of all classes and instances, $\mathcal{R}$ is the set of all properties, while $\mathcal{L}$ represents the set of all literal values.}
where $sb$ represents a subject (a class or an instance), 
$p$ represents a predicate (a property) 
and $ob$ represents an object
(a class, an instance or a literal). 
%
KG entities (\ie $\mathcal{E}\cup\mathcal{R}$: classes, properties and instances) are represented by an URI (Uniform Resource Identifier).

An (ontology-enhanced) KG can be split into a TBox (terminology) and an ABox (assertions). The TBox is composed by triples 
using  RDF  Schema (RDFS) constructors  like  class  subsumptions \linebreak 
and property domain 
and range;
and OWL constructors like disjointness, equivalence and property inverses.\footnote{Note that the Web Ontology Language (OWL) \cite{owl2} also enables the creation of complex axioms that are translated/serialized into more than one triple: \url{https://www.w3.org/TR/owl2-mapping-to-rdf/}} 
The ABox contains assertions 
among instances, including OWL equality and inequality,
and semantic type definitions. 
Table \ref{tab:triples} shows several examples of TBox and ABox triples.



\subsection{Ontology alignment}
Ontology alignment is the process of finding mappings or
correspondences between a source and a target ontology or knowledge graph \cite{om2013,om-tkde2013}.
These mappings typically represent  equivalences or broader/narrower relationships among the entities of
the input ontologies. 
In the ontology matching community \cite{oaei2020}, mappings are exchanged using the RDF Alignment format \cite{alignAPI2011}; but they can also be interpreted as standard OWL axioms (\eg \cite{jbs2011,bioportalrepair2014}). In this work we treat ontology alignments as OWL axioms (\eg Triple \ref{ax:et-ncbi} in Table \ref{tab:triples}). 
%
An ontology matching system (\eg LogMap~\cite{logmap2011}) is a program
that, given as input two ontologies or knowledge graphs,
generates as output a set of mappings (\ie an alignment) $M$.

\subsection{Embedding models}
Knowledge graph embedding (KGE)~\cite{KGE_survey_2017,DBLP:journals/tkdd/RossiBFMM21} plays a key role in link prediction problems where it is applied to knowledge graphs to resolve missing facts in largely connected knowledge graphs, such as DBpedia  \cite{dbpedia2015}. Biomedical link prediction is another area where embedding models have been applied successfully
(\eg~\cite{DBLP:journals/bioinformatics/AlshahraniKMKQH17,DBLP:journals/ws/AgibetovS20}).

The embeddings of the entities in a KG are commonly learned by \textit{(i)} defining a scoring function over a triple, which is typically proportional to the  probability of the existence of that triple in the KG,\footnote{For the embedding process, we focus on triples where $o\in\mathcal{E}$ is a class or an instance.} \ie $SF : \mathcal{E}\times\mathcal{R}\times\mathcal{E}\to \mathbb{R}$, $SF \propto P(\left\langle sb, p, ob \right\rangle\in KG)$; and \textit{(ii)}~minimizing a loss function (\ie deviation of the prediction of the scoring function with respect to the truth available in the KG). More specifically, KGE models \textit{(i)} initialize the entities in a triple $\left\langle sb, p, ob \right\rangle$ into a vector representation $\vec{e}_{sb}, \vec{e}_p,\vec{e}_{ob} \in \mathbb{R}^{k} \text{~or~} \mathbb{C}^{k}$, where $k$ is the dimension of the vector; \textit{(ii)} apply a scoring function to $(\vec{e}_{sb}, \vec{e}_p, \vec{e}_{ob})$; and \textit{(iii)} adapt the vector representations to improve the scoring and minimize the loss.


Several knowledge graph embedding models have been proposed. In this work, we used models of three major categories: decomposition models, geometric models, and convolutional models.\footnote{The interested reader please refer to \cite{DBLP:journals/tkdd/RossiBFMM21} for a comprehensive survey.}
The decomposition models 
represent the triples of the KG into a one-hot
3-order tensor and apply matrix decomposition to
learn entity vectors.
%
%
Geometric models, also known as translational, try to
learn embeddings by defining a scoring function where the predicate in the triple act as a geometric translation (\eg rotation)
from subject to object. 
%
Convolutional models, unlike previous models, learn entity embedding with
non-linear scoring functions via convolutional 
layers.
%
}

\section{Ecotoxicological risk assessment and adverse biological effect prediction}
\label{sec:niva}


The task of ecotoxicological risk assessment is to study the potential hazardous effects of chemicals on organisms from individuals to ecosystems. In this context, risk is the result of the intrinsic hazards of a substance on species, populations or ecosystems, combined with an estimate of the environmental exposure, \ie the product of exposure and effect (hazard).

\begin{figure}[h]
\centering
\includegraphics[width=0.47\textwidth]{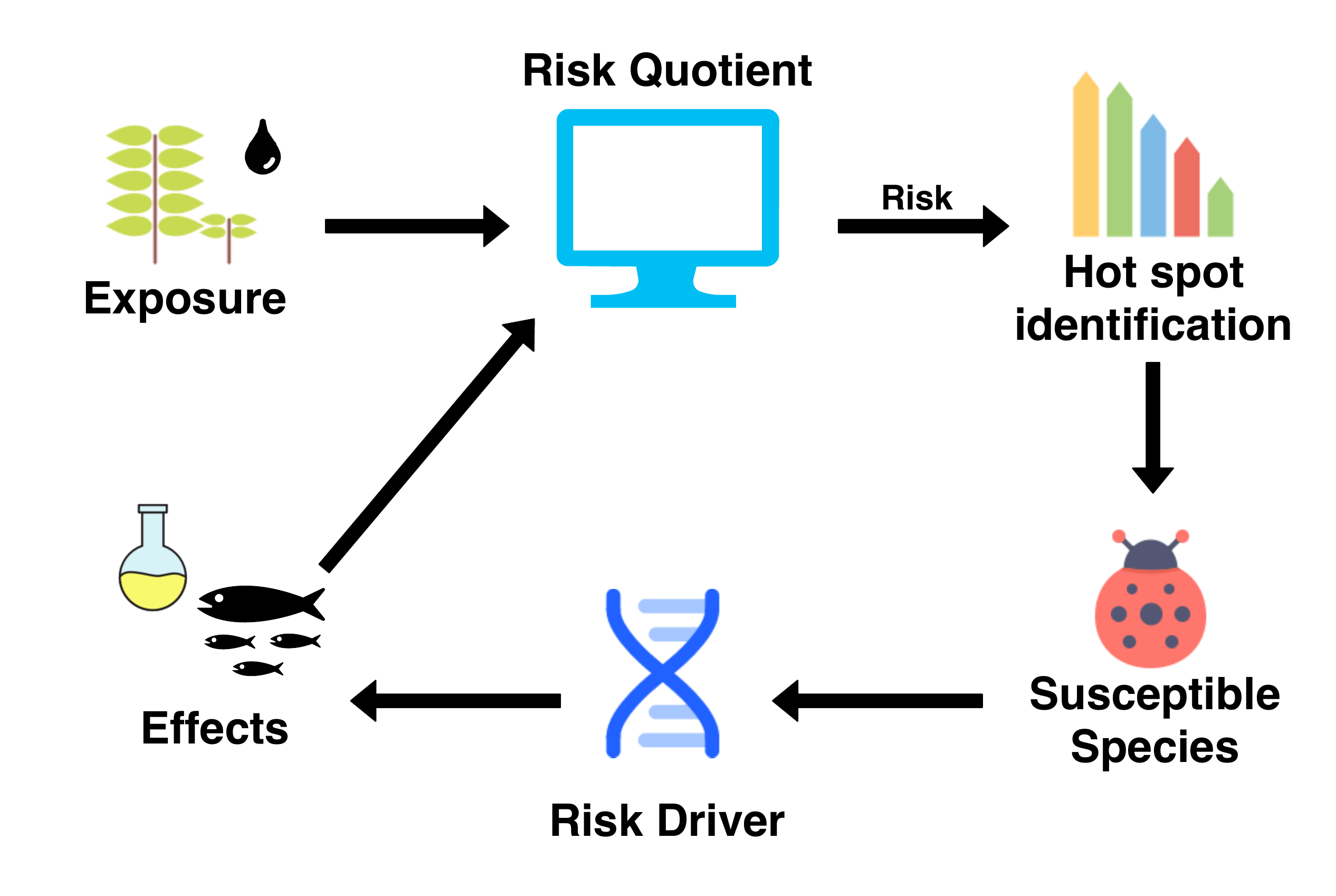}
\caption{Simplified ecological risk assessment pipeline.}
\label{fig:niva-pipeline}

\end{figure}

Figure \ref{fig:niva-pipeline} shows a simplified risk assessment pipeline.  %
\textit{Exposure} data is gathered from analysis of environmental concentrations of one or more chemicals, while \textit{effects} (\textit{hazards}) are characterized for a number of species in the laboratory as a proxy for more ecologically relevant organisms. These two data sources are used to calculate the so-called risk quotient (RQ; ratio between exposure and effects). The RQ for one chemical or the mixture of many chemicals is used to identify chemicals with the highest RQs (risk drivers), identify relevant modes of action\footnote{The mode of action describes the molecular pathway by which a chemical causes physiological change in an organism.} (MoA) and characterize detailed toxicity mechanisms for one or more species (or taxa). Results from these predictions can generate a number of new hypotheses that can be investigated in the laboratory or studied in the environment. 
\change{Note that, this risk assessment pipeline is a simplified version of the one in use at the Norwegian Institute for Water Research,\footnote{NIVA: \url{https://www.niva.no/en}} however, similar methodologies are used across regulatory risk assessment pipelines.}

\begin{table*}[tb]
    \centering
    \begin{tabular}{|c|c|l|}
    \hline
        \textbf{Endpoint} & \textbf{Frequency} & \textbf{Description} \\ \hline
        NR & $0.21$ & Not reported\\\hline
        NOEL & $0.17$ & No-observable-effect-level\\\hline
        LC50 & $0.16$ & Lethal concentration for $50\%$ of test population\\\hline
        LOEL & $0.14$ & Lowest-observable-effect-level\\\hline
        NOEC & $0.05$ &No-observable-effect-concentration\\\hline
        EC50 & $0.05$ &Effective concentration for $50\%$ of test population\\\hline
        LOEC & $0.04$ &Lowest observable effect concentration\\\hline
        BCF & $0.03$ & Bioconcentration factor\\\hline
        NR-LETH & $0.02$ &Lethal to $100\%$ of test population\\\hline
        LD50 & $0.02$ & Lethal dose for $50\%$ of test population \\\hline
        Other & $0.11$ & \\\hline
    \end{tabular}
    \medskip
    \caption{The most frequent endpoints in ECOTOX \cite{ecotox} chemical effect data.}
    \label{tab:endpoints}
    
\end{table*}

The chemical effect data is gathered during laboratory experiments, where a sub-population of a single species is exposed to an increasing concentration of a toxic chemical. 
The \textit{endpoints} of the experiments are recorded at chemical concentrations and time after exposure. 
These \textit{endpoints} are categorized into several categories, \eg lethality rate of test population (see Table \ref{tab:endpoints}).

%

Ecological risk assessment methods require a large amount of these experimental data to give an accurate depiction of the long term risk to an ecosystem. The data must cover the relevant chemicals and species present in the ecosystem, \eg an ecological risk assessment of agricultural runoff in Norway will mostly concern pesticides and waterflees, copepods, and frogs, among other species \cite{smadyr}. Just with a few relevant chemicals and species the search space becomes immense and performing laboratory experiments becomes unfeasible. Thus, it is essential to develop \textit{in~silico} methods to extrapolate new chemical-species effects from known combinations. We differentiate among two types 
complementary strategies:
\begin{inparaenum}[\it (i)]
\item highly specialized (restricted in chemical and species domains) models to predict chemical concentrations that will have an effect on a test species, and 
\item models that produce rankings of highly representative chemical-species pair hypothesis which can be used by a laboratory to perform targeted experiments. 
\end{inparaenum}
In this paper we focus on the latter strategy, using a method based on knowledge graph embeddings. 
Methods that fall into the first strategy are introduced in Section \ref{sec:rel-tox}.

\section{Related work}
\label{sec:related}

This section will cover related work from ecotoxicology and knowledge graph based prediction.

\subsection{Toxicity extrapolation}
\label{sec:rel-tox}

There are two main research areas in toxicology to extrapolate chemical effects,
\ie Quantitative Structure-Activity Relationship (QSAR) and read-across. QSAR
modelling try to find a relationship between the structure of a chemical and the chemical's biological activity (\cf reviews \cite{Arkadiusz2006,10.1093/mutage/gey046}). This relationship is described using derived chemical features. Some features are simple, \eg octanol-water partition coefficient or logP, others concern the entire chemical, \eg chemical fingerprints. The basis of the QSAR relationship is usually modeled as polynomial equations. Parthasarathi and Dhawan \cite{PARTHASARATHI201891} take
this further by using the logarithm of chemical concentration to achieve a polynomial relationship: $\log(1/\kappa)=f(\pi)+g(\sigma)$, $f\in P_2$ and $g\in P_1$ ($P_n$ is a polynomial of $n$th degree), where
$\kappa$ is the chemical concentration while $\pi$ and $\sigma$ denote the derived chemical features hydrophobicity\footnote{Measure of the absence of attraction to water.} and electronic effects in the molecule, respectively. The drawback of these models is the applicability domains. Usually, a QSAR model considers a small set of chemicals (10ths to 100ths) and one single species. 
This means that new features and relationships 
need to be developed for each species and each chemical~group. 

The read-across methods try to mitigate these drawbacks, mainly by considering extrapolation of the effect at the chemical and species levels. Similar to QSAR models, read-across of chemicals use the chemical features to create similarity measures between chemicals to justify the read-across of chemical effects. The read-across in the species domain is harder. Species do not tend to have easily derived features. 
Therefore, genetic similarity has emerged as a viable option. Sequence Alignment to Predict Across Species Susceptibility (SeqAPASS), developed by the United States Environmental Protection Agency (U.S. EPA.), is an example of such an approach~\cite{seqapass,10.1093/toxsci/kfy186}. 
SeqAPASS uses a large amount of data available for humans, mice, rats, and zebrafish to extrapolate to areas with lower coverage.


\subsection{Embedding models}
\label{sec:relatedKGEM}

In this work, we use nine KGE models across three categories of models. Here, we will give a brief introduction to the models, while a more extended explanation of the models is found in Appendix \ref{sec:appendix_kge}. The interested reader please refer to \cite{DBLP:journals/tkdd/RossiBFMM21} for a comprehensive survey.

The three categories of models are decomposition, geometric, and convolutional \cite{DBLP:journals/tkdd/RossiBFMM21}. The decomposition models are DistMult, ComplEx, and HolE.
DistMult models the score of a triple as the vector multiplication of the representation of each subject, predicate and object \cite{Yang2015EmbeddingEA}.
ComplEx 
uses the same scoring function as DistMult, however, in a complex vector space, such that it can handle inverse relations \cite{DBLP:journals/corr/TrouillonWRGB16}. HolE is based on holographic embeddings \cite{DBLP:journals/corr/NickelRP15}, however, it has been shown that HolE is equivalent to ComplEx \cite{hayashi-shimbo-2017-equivalence}.

The geometric models are TransE, RotatE, pRotatE, and HAKE. TransE is the base of a whole family of models and scores triples based on the translation from subject to object using the representation of the predicate \cite{NIPS2013_5071}. RotatE is similar to TransE, however, the translation using the predicate is done by rotating it (via Euler's identity) \cite{sun2018rotate}. Furthermore, pRotatE is a baseline for RotatE where the modulus in Euler's identity is ignored \cite{sun2018rotate}. Finally, the hierarchical-aware model, HAKE, where entities at each level in the hierarchy is at equal distance from the origin and relations at a level is modeled as rotation \cite{zhang2019learning}. 

The convolutional models take a deep learning approach to the task of KGE. We use ConvKB \cite{Nguyen2018} and ConvE \cite{dettmers2018}, which are similar with slightly different architectures. They have shown good performance given the relative small number of parameters. 

\change{
Although quite a few KGE models have been proposed, the adopted ones are either classic models or can achieve state-of-the-art performance in some benchmarks.
They are representative of mainstream techniques, and have been widely adopted in KGE research and applications \cite{DBLP:journals/tkdd/RossiBFMM21}.
Thus, the benefits and shortcomings of the KGE models analysed in this study 
provide good evidence of the general performance of this type of models in a complex prediction task, \ie adverse biological effect of chemicals on organisms.
}

\subsection{Using KGE for prediction}

Our focus to use KGE models is to predict if a chemical has a lethal effect on an organism.
KGE models have been explored in the biomedical domain to solve similar predictions tasks (\eg finding relationships between diseases, drugs, genes, and treatments). 
%
Several works have shown improvements in results by using KGE models for prediction, \eg \cite{10.1371/journal.pone.0218264,DBLP:journals/bioinformatics/AlshahraniKMKQH17,DBLP:journals/ws/AgibetovS20}. Chen~et~al.~\cite{C2MB00002D} used random walks over networks to perform drug-target predictions. The ChEMBL and DrugBank KGs have also been used to predict chemical mode of action (MoA) of anticancer drugs with high performance on benchmark datasets~\cite{doi:10.1111/bph.13629}. 

Opa2vec \cite{smaili2019opa2vec} and Blagec et al. \cite{Blagec2019} have developed embedding models to improve similarity-based prediction in the biomedical domain, while OpenBioLink~\cite{10.1093/bioinformatics/btaa274} 
has created a framework for evaluating models in the biomedical domain. 

EL Embeddings \cite{kulmanov2019embeddings} and opa2vec \cite{smaili2019opa2vec} present new semantic embedding methods for KGs with expressive logic expressions (\ie OWL ontologies) to predict protein interaction.
The former utilizes complex geometric structures to model the logic relationships between entities, while the later learns a language model from a corpus extracted from the ontology. OWL2Vec*~\cite{owl2vec2020} also learns a language model from an ontology and applies the computed embeddings into two prediction tasks: class subsumption and class membership. OWL2Vec* has also been used to predict the plausibility of ontology alignments \cite{owl2vecalignment}.


\change{
To the best of our knowledge there is no work using link prediction or KGE models to support 
ecotoxicological effect prediction. 
This study will give novel insights and empirical results of KGE models 
in this new~domain.
}



\section{TERA knowledge graph}
\label{sec:tera}

\begin{figure*}[t]
     \centering
     \includegraphics[width=0.95\textwidth]{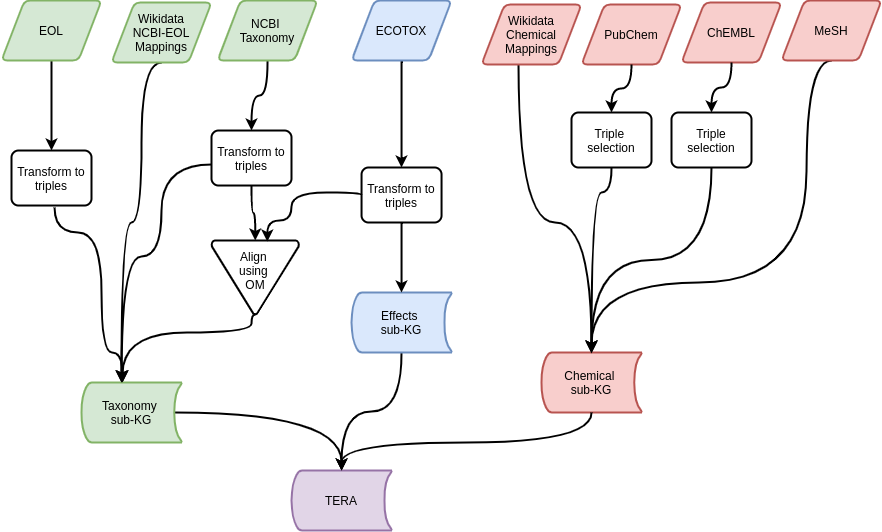}
     \caption{Data sources and processes to create the TERA knowledge graph.}
     \label{fig:dataflow}
\end{figure*}

One major challenge in ecological risk assessment processes is the interoperability of data. 
In this section, we introduce the Toxicological Effect and Risk Assessment (TERA), an ontology-enhanced RDF-based knowledge graph that aims at providing an integrated view of the relevant data sources for risk assessment.\footnote{Resources to create and access TERA: \url{https://github.com/NIVA-Knowledge-Graph/TERA}}

\change{The initial inspiration for TERA was the aid of ecotoxicological effect prediction where access to disparate resources was required (see Section \ref{sec:dataaccess}).
However, by integrating these sources into a KG, we were also able to directly apply TERA into the prediction process by leveraging knowledge graph embedding models (see Section \ref{sec:teraprediction}).}

\change{The data sources integrated into TERA vary from tabular and RDF files to SPARQL endpoints over public linked data. The sources currently integrated into TERA are: \begin{inparaenum}
[\it (i)]
\item biological: NCBI Taxonomy, Encyclopedia of Life, and Wikidata mappings ($\sim 500k$ species);
\item chemical: PubChem, ChEMBL, MeSH, and Wikidata mappings ($\sim 110M$ compounds); and
\item biological effects: ECOTOXicology Knowledgebase ($\sim 1M$ results, $\sim 12k$ compounds, $ \sim 13k$ species), and system-generated mappings.
\end{inparaenum}
These three distinct parts make up the sub-KGs of TERA, \ie \begin{inparaenum}[\it (i)]
\item the Taxonomy sub-KG (KG$_S$),
\item the Chemical sub-KG (KG$_C$), and 
\item the Effects sub-KG (KG$_E$).
\end{inparaenum}
The different processes to transform and integrate these sources into TERA are shown in Figure \ref{fig:dataflow}.
}

%

A snapshot of TERA is available on Zenodo \cite{myklebust_erik_bryhn_2019_4244313}, where licenses permit.\footnote{
EOL: Various Creative commons (CC),
NCBI: Creative Commons CC0 1.0 Universal (CC0 1.0),
ECOTOX: No restrictions,
PubChem: Open Data Commons Open Database License, 
ChEMBL:  CC Attribution,
MeSH: Open, \emph{Courtesy of the U.S. National Library of Medicine},
Wikidata: CC0 1.0.}
PubChem and ChEMBL 
are not included in the snapshot due to size constraints; these can be downloaded from the National Institutes of Health\footnote{\url{ftp://ftp.ncbi.nlm.nih.gov/pubchem/RDF/}} and European Bioinformatics Institute,\footnote{\url{ftp://ftp.ebi.ac.uk/pub/databases/chembl/ChEMBL-RDF/}} respectively.
The subgraph of TERA used for prediction is available alongside the chemical effect prediction models in our GitHub repository.\footnote{\url{https://github.com/NIVA-Knowledge-Graph/KGs_and_Effect_Prediction_2020}}  
%
Table \ref{tab:triples} shows several examples of RDF triples from TERA.\footnote{Prefixes associated to the URI namespaces of entities in TERA: 
\texttt{et:} (ECOTOXicology knowledgebase),
\texttt{ncbi:} (NCBI taxonomy), 
\texttt{eol:} (Encyclopedia of Life), \texttt{mesh:} (Medical Subject Heading), \texttt{compound:} (PubChem compound), \texttt{descr:} (PubChem descriptors), \texttt{vocab:} (PubChem vocabulary), 
\texttt{inchikey:} (InChIKey identifiers), 
\texttt{envo:} (Environment Ontology)
\texttt{cheminf:} (Chemical information ontology),
\texttt{chembl:} (ChEMBL), 
\texttt{chembl$\_$m:} (ChEMBL molecule subset),
\texttt{chembl$\_$t:} (ChEMBL target subset),
\texttt{wd:} (WikiData entities), 
\texttt{wdt:} (Wikidata properties), 
\texttt{qudt:} (Quantities, Units, Dimensions and Types Catalog),
\texttt{snomedct:} (SNOMED CT ontology), and
\texttt{bp:} (Biological PAthway eXchange ontology).
\texttt{owl:}, \texttt{rdfs:}, \texttt{rdf:} and \texttt{xsd:} 
are prefixes referring to W3C standard vocabularies.}


%


\begin{table*}[t]
    \centering
    \begin{tabular}{|c|c|c|c|c|}
        \hline
        test\_id & reference\_number & test\_cas & species\_number & organism\_habitat \\ \hline 
        $1147366$ & $12448$ & $134623$ (diethyltoluamide) & $1$ (\emph{Pimephales promelas}) & Water \\
        \hline
    \end{tabular}
    \medskip
    \caption{ECOTOX database tests example.}
    \label{tab:ecotox_ex1}
\end{table*}

\begin{table*}[t]
    \centering
    \begin{tabular}{|c|c|c|c|c|c|}
        \hline
        result\_id & test\_id & endpoint & effect & conc1\_mean & conc1\_unit \\ \hline 
        $102570$ & $1147366$ & $LC50$ & MOR & $110000$ & $\mu g/L$ \\
        \hline
    \end{tabular}
    \medskip
    \caption{ECOTOX database results example.}
    \label{tab:ecotox_ex2}
\end{table*}

\subsection{Dataset overview}

TERA, as mentioned above, is constructed by gathering a number of sources about chemicals, species and chemical toxicity, with a diverse set of formats including tabular data, RDF dumps and SPARQL~endpoints.

\medskip
\noindent
\textit{Biological effect data of chemicals.} 
The largest publicly available repository of effect data is the ECOTOXicology knowledgebase (ECOTOX) developed by the US Environmental Protection Agency~\cite{ecotox}. This data is gathered from published toxicological studies and limited internal experiments.
The dataset consists of $1M$ experiments covering $12k$ chemicals and $13k$ species,\footnote{Version dated Sep. 15, 2020.} implying a chemical--species pair converge of maximum $\sim 0.6\%$. The resulting endpoint from an experiment is categorised in one of a plethora of predefined endpoints (see Table \ref{tab:endpoints} above).


Tables \ref{tab:ecotox_ex1} and \ref{tab:ecotox_ex2}
contain an excerpt of the ECOTOX database. 
ECOTOX includes information about the chemicals and species used in the tests. 
This information, however, is limited and additional (external) resources are required to complement ECOTOX.

\medskip
\noindent
\textit{Chemicals.}
The ECOTOX database 
uses an identifier called CAS Registry Number assigned by the Chemical Abstracts Service to identify chemicals. The CAS numbers are proprietary, however, Wikidata \cite{wikidata2014} (indirectly) encodes mappings between CAS numbers and open identifiers like \textit{InChIKey}, a 27-character hash of the International Chemical Identifier (InChI) which encodes chemical information uniquely \cite{Heller2015}.\footnote{While InChI is unique, InChiKey is not, and collisions have greater than zero probability \cite{Willighagen2011}.}
Wikidata also provides mappings to well known databases like PubChem, ChEMBL and MeSH, which include relevant chemical
information such as chemical structure, structural classification and functional classification.

\medskip
\noindent
\textit{Taxonomy.}
ECOTOX contains a taxonomy\footnote{In the context of the paper ``taxonomy'' typically refers to a classification of organisms.} (of species),
however, this only considers the species represented in the ECOTOX effect data. Hence, to enable extrapolation of effects across a larger taxonomic domain, we include the NCBI Taxonomy~\cite{ncbi}. This taxonomy data source
consists
of a number of database dump files, which contains a hierarchy for all sequenced species, which equates to around $10\%$ of the currently known life on Earth and \change{is one of the most comprehensive taxonomic resources}. For each of the taxa (species and classes), the taxonomy defines a handful of labels, 
the most commonly used of which are the \emph{scientific} and \emph{common} names. However, labels such as \emph{authority} can be used to see the citation where the species was first mentioned, while \emph{synonym} is a alternate \emph{scientific} name, that may be used in the literature. 

\medskip
\noindent
\textit{Species traits.}
As an analog to chemical features, we use species traits to expand the coverage of the knowledge graph. \change{Apart from taxonomic classifications, traits are the most important 
information to identify species and will be of great importance when predicting the effect on the species.}

The traits we have included in the knowledge graph are the habitat, endemic regions, and presence (and classifications of these). This data is gathered from the Encyclopedia of Life (EOL) \cite{eol}, which is available as a property graph. Moreover, EOL uses external definitions of certain concepts, and mappings to these sources are available as glossary files.
In addition to traits, researchers may be interested in species that have different conservation statuses, \eg if the population is stable or declining, etc. This data can also be extracted from EOL.


\subsection{Dataset preprocessing}
\label{sec:datapreparation}

In this section we present the different steps to extract, transform and integrate the source datasets into the main TERA components and sub-KGs. \change{All data is transformed using custom mappings (scripts) from the sources to RDF triples.}
Table~\ref{tab:triples} shows an excerpt of the triples in TERA.


\setlength{\tabcolsep}{3.0pt}
\begin{table*}[p!]
    \centering
    \begin{tabular}{|N c c c|}
        \hline
        \multicolumn{1}{|c}{\texttt{\#}} & \texttt{subject} & \texttt{predicate} & \texttt{object} \\ \hline\hline
        
        \multicolumn{4}{|c|}{Effects sub-KG}\\\hline\hline
        
        \label{ax:test1} & \texttt{et:test/1147366} & \texttt{et:compound} & \texttt{et:chemical/134623} \\
        \label{ax:test2} & \texttt{et:test/1147366} & \texttt{et:species} & \texttt{et:taxon/1} \\

        \label{ax:result} & \texttt{et:test/1147366} & \texttt{et:hasResult} & \texttt{et:result/102570} \\
        
        \label{ax:result2} & \texttt{et:result/102570} & \texttt{et:endpoint} & \texttt{et:endpoint/LC50}\\
        \label{ax:result3} & \texttt{et:result/102570} & \texttt{et:effect} & \texttt{et:effect/Mortality}\\

        \label{ax:et-tax1} & \texttt{et:taxon/1} & \texttt{rdf:type} & \texttt{et:taxon/Pimephales} \\
        
        \label{ax:et-tax2} & \texttt{et:taxon/Pimephales} & \texttt{rdfs:subClassOf} & \texttt{et:taxon/Cyprinidae} \\
        
        \label{ax:et-label1} & \texttt{et:taxon/1} & \texttt{et:latinName} & \texttt{``Pimephales promelas''} \\
        
        \label{ax:et-label2} & \texttt{et:taxon/1} & \texttt{et:commonName} & \texttt{``Fathead Minnow''} \\

        \label{ax:et-group} & \texttt{et:taxon/1} & \texttt{et:speciesGroup} & \texttt{et:group/Fish} \\
        
         \label{ax:et-rank} & \texttt{et:taxon/1} & \texttt{et:rank} & \texttt{et:rank/species} \\
        
        \label{ax:et-label3} & \texttt{et:chemical/134623} & \texttt{rdfs:label} & \texttt{``diethyltoluamide''} \\

        \hline\hline

        \multicolumn{4}{|c|}{Entity Mappings}\\\hline\hline
        
        \label{ax:et-ncbi} & \texttt{et:taxon/1} & \texttt{owl:sameAs} & \texttt{ncbi:taxon/90988} \\\hline
        \label{ax:ncbi-wd} & \texttt{ncbi:taxon/90988} & \texttt{owl:sameAs} & \texttt{wd:Q2700010} \\
         
         \label{ax:wd-eol}& \texttt{wd:Q2700010}  & \texttt{owl:sameAs} & \texttt{eol:211492} \\\hline 
        
        \label{ax:et-wd}& \texttt{et:chemical/134623} & \texttt{owl:sameAs} & \texttt{wd:Q408389} \\
         
         \label{ax:wd-chembl}& \texttt{wd:Q408389}  & \texttt{owl:sameAs} & \texttt{chembl$\_$m:CHEMBL1453317} \\
         \label{ax:wd-pubchem}& \texttt{wd:Q408389}  & \texttt{owl:sameAs} & \texttt{compound:CID4284} \\
         \label{ax:wd-mesh}& \texttt{wd:Q408389} & \texttt{owl:sameAs} & \texttt{mesh:D003671} \\
         \label{ax:wd-inchikey} & \texttt{wd:Q408389}  & \texttt{owl:sameAs} & \texttt{inchikey:MMOXZBCLC\ldots}$^1$ \\
        \hline

        \multicolumn{4}{|c|}{Taxonomy sub-KG}\\\hline\hline
        


        \label{ax:tax-type1} & \texttt{ncbi:taxon/90988} & \texttt{rdf:type} & \texttt{ncbi:taxon/51137}$^2$ \\
       \label{ax:tax-type2} & \texttt{ncbi:taxon/90988} & \texttt{rdf:type} & \texttt{ncbi:division/10} \\
        
        \label{ax:name} & \texttt{ncbi:taxon/90988} & \texttt{ncbi:scientific$\_$name} & \texttt{``Pimephales promelas''} \\

        \label{ax:rank} & \texttt{ncbi:taxon/90988} & \texttt{ncbi:rank} & \texttt{ncbi:species} \\
        
        \label{ax:sub} & \texttt{ncbi:taxon/51137} & \texttt{rdfs:subClassOf} & \texttt{ncbi:taxon/7953}  $^3$ \\

        \label{ax:label} & \texttt{ncbi:division/10} & \texttt{rdfs:label} & \texttt{``Vertebrates''} \\
        
        \label{ax:disj} & \texttt{ncbi:division/10} & \texttt{owl:disjointWith} & \texttt{ncbi:division/1}   \\
        
        \label{ax:label2} & \texttt{ncbi:division/1} & \texttt{rdfs:label} & \texttt{``Invertebrates''} \\

        
        \label{ax:habitat} & \texttt{eol:211492} & \texttt{eol:habitat} & \texttt{envo:00000153} $^4$  \\\hline\hline
   
   

        \multicolumn{4}{|c|}{Chemical sub-KG}\\\hline\hline

      \label{ax:mesh_hier} & \texttt{mesh:D003671} & \texttt{mesh:broaderDescriptor} & \texttt{mesh:D001549} $^5$ \\
    \label{ax:mesh_hier_func} & \texttt{mesh:D003671} & \texttt{mesh:pharmacologicalAction} & \texttt{mesh:D007302} $^6$ \\ 
    \label{ax:has_target} & \texttt{chembl$\_$m:CHEMBL1453317} & \texttt{chembl:hasTarget} & \texttt{chembl$\_$t:CHEMBL1907594} $^7$ \\
    \label{ax:has_target_rel} & \texttt{chembl$\_$t:CHEMBL1907594} & \texttt{chembl:relSubsetOf} & \texttt{chembl$\_$t:CHEMBL3137273} $^8$ \\
    
    \label{ax:parent_compound} & \texttt{compound:CID89845769} $^9$ & \texttt{vocab:hasParentCompound} & \texttt{compound:CID4284} \\
    \label{ax:has_component} & \texttt{compound:CID131721069} $^{10}$ &  \texttt{cheminf:CHEMINF$\_$000478} $^{11}$  & \texttt{compound:CID4284} \\
    \label{ax:small_molecule} & \texttt{compound:CID131721069} & \texttt{rdf:type} &  \texttt{bp:SmallMolecule} \\
    \label{ax:is_active} & \texttt{compound:CID7547} $^{12}$ &  \texttt{~~vocab:is$\_$active$\_$ingredient$\_$of~~} & \texttt{snomedct:411346009} $^{13}$
    \\
    \label{ax:similar_to} & \texttt{compound:CID131721069} &  \texttt{cheminf:CHEMINF$\_$000480} $^{14}$  & \texttt{compound:CID10751691} $^{15}$
    \\

    
    
    \hline
    
    \end{tabular}
    \smallskip
    \caption{Example triples from the TERA knowledge graph. For space reasons, we have added the full id or label for some of the entities using footnote marks where
    $^1$inchikey:MMOXZBCLCQITDF-UHFFFAOYSA-N,
    $^2$Pimephales,
    $^3$Cyprinidae,
    $^4$Headwater,
    $^5$Benzamides,
    $^6$Insect Repellents,
    $^7$CHRNA3,
    $^8$CHRNB4,
    $^9$DETA-20,
    $^{10}$DETA Epichlorohydrin, 
    $^{11}$Has component, 
    $^{12}$Triclocarban,
    $^{13}$Trichlorocarbanilide-containing product, 
    $^{14}$Similar to,
    $^{15}$3-Chloromethyl-N,N-diethylbenzamide. 
    }
    \label{tab:triples}
\end{table*}


 \begin{figure*}[t!]
    \centering
    \includegraphics[width=0.99\textwidth]{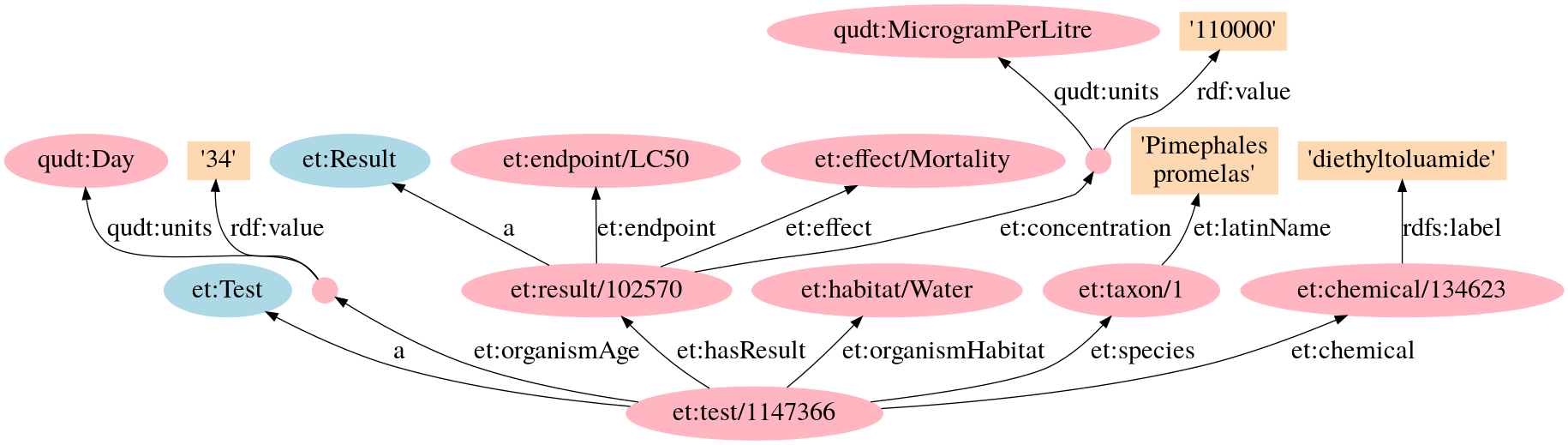}
    \caption{Example of an ECOTOX test and related triples.}
    \label{fig:ecotox_kg_ex}
\end{figure*}

\subsubsection{Effects sub-KG construction}

The effect data in ECOTOX consist of two parts, \ie test definitions and results associated with the test definitions (see Tables \ref{tab:ecotox_ex1} and \ref{tab:ecotox_ex2}, respectively).
%
The important 
columns of a test are the chemical and the species used.
Other columns include metadata, but these are optional and often empty. Each result is composed by an endpoint, an effect, and a concentration 
(with a unit) at which the endpoint and effect 
are recorded.

This tabular data in ECOTOX  is transformed into triples that form the \textit{effects sub-KG} in TERA ($KG_E$). Note that a test can have multiple results. 
A subset of the effect triples are listed in Table \ref{tab:triples} (see Triples \texttt{\ref{ax:test1}-\ref{ax:et-label3}}). A graphical representation for an effect test and its result is also shown in Figure \ref{fig:ecotox_kg_ex}.


 ECOTOX contains metadata about the species and chemicals used in the experiments. This metadata is also included 
 in TERA to facilitate the alignment with other resources (see Section \ref{sec:alignment}). 
  \begin{enumerate}[\it (i)]
     \item The ECOTOX metadata file  \emph{species.txt} 
     includes common and Latin names, along with a (species) ECOTOX group (see triples \texttt{\ref{ax:et-label1}}-\texttt{\ref{ax:et-group}} in Table \ref{tab:triples}). This group is a categorization of the species based on ECOTOX use cases. Prefixes and abbreviations like \emph{sp.}, \emph{var.} are removed from the label names.
    \item The full hierarchical lineage\footnote{As defined by U.S. EPA. Note that species hierarchies are contested among researchers.} is also available in the metadata file \emph{species.txt}.
        Each column 
        represents a taxonomic level, \eg \emph{genus} or \emph{family}. If a column is empty, we construct 
        an intermediate classification; for example, 
        \emph{Daphnia magna} has no genus classification in the data, then its classification 
        is set to Daphniidae genus (family name + genus, actually called \emph{Daphnia}). We construct these classifications to ensure the number of levels in the taxonomy is consistent (see triples \texttt{\ref{ax:et-tax1}} and \texttt{\ref{ax:et-tax2}} in Table \ref{tab:triples}). 
        Note that when adding triples such as \texttt{\ref{ax:et-rank}} in Table \ref{tab:triples}, we also add a taxonomic rank to facilitate the querying for a specific taxonomic level.

        \item The ECOTOX source file \emph{chemicals.txt} includes chemical metadata and it is handled similarly to \emph{species.txt}. The file includes chemical name (see~\texttt{\ref{ax:et-label3}} in Table \ref{tab:triples}) and a (chemical) ECOTOX group. 
    \end{enumerate}

For the units in the effect data, \eg chemical concentrations (mg/L, mol/L, mg/kg, etc.), we reuse the \texttt{QUDT}~1.1\footnote{QUDT 1.1: \url{http://linkedmodel.org/catalog/qudt/1.1/}} ontologies. 
When an unit such as mg/L is not defined, we define it 
according to Listing \ref{lst:unit_example}. 

\begin{code*}[t!]
\begin{minted}{emacs}
@prefix rdf:     <http://www.w3.org/1999/02/22-rdf-syntax-ns#> .
@prefix rdfs:    <http://www.w3.org/2000/01/rdf-schema#> .
@prefix qudt: <http://qudt.org/schema/qudt#> .
@prefix et:   <https://cfpub.epa.gov/ecotox> . 
et:MilligramPerLiter
    rdf:type qudt:MassPerVolumeUnit, qudt:SIDerivedUnit ;
    rdfs:label "Milligram per Liter"^^xsd:string ;
    qudt:abbreviation "mg/L"^^xsd:string ;
    qudt:conversionMultiplier 0.000001 ;
    qudt:conversionOffset 0.0 ;
    qudt:symbol "mg/dm^3"^^xsd:string .
\end{minted}
\vspace{-0.5cm}
\caption{Unit definition of mg/L using \texttt{QUDT}.}
\label{lst:unit_example}
\end{code*}

\subsubsection{Alignment with state-of-the-art tools}
\label{sec:alignment}

ECOTOX database provides proprietary chemical identifiers (\ie CAS numbers) and internal ECOTOX ids for species. In order to extrapolate effects across a larger set of chemicals and species than those available in ECOTOX, 
TERA integrates taxonomy and trait data from NCBI and EOL, and chemical data from PubChem, ChEMBL and MeSH

\smallskip
\noindent
\textit{Alignment 
between ECOTOX and the NCBI Taxonomy}.
There does not exist a complete and public alignment between the 23,439 ECOTOX species and the 1,830,312 the NCBI Taxonomy species.\footnote{There 
are a total of 27,133 and 2,246,074 taxa in ECOTOX and NCBI, respectively. However, we focus on species, \ie instances.} We have used three methods, two state-of-art ontology alignments systems and a baseline, to align ECOTOX and the NCBI Taxonomy: 
\begin{inparaenum}[\it (i)]
\item LogMap  \cite{logmap2011,logma_ecai2012},
\item AgreementMakerLight (AML) \cite{10.1007/978-3-642-41030-7_38}, and 
\item a string matching algorithm based on Levenshtein distance \cite{1966SPhD...10..707L}.
\end{inparaenum} 
\change{LogMap and AML were chosen since they have performed well across many datasets in the Ontology Alignment Evaluation Initiative (\eg  \cite{oaei2018,oaei2019,oaei2020}). Most mappings in our setting are expected to be lexical, therefore, we also selected a purely lexical matcher to evaluate if more sophisticated systems like LogMap and AML bring an additional value.}

\change{Due to the large size of the NCBI Taxonomy, we needed to split NCBI into manageable chunks to enable the use of ontology alignment systems. Fortunately, this can be easily done by considering the species division, \eg mammal or invertebrate. This divides the NCBI Taxonomy into 11 distinct parts, which can be aligned to the taxonomy in ECOTOX.}

\setlength{\tabcolsep}{5.5pt}
\begin{table}[t!]
    \centering
    \vspace{0.5cm}
    \begin{tabular}{|l||c|c|c|}
    \hline 
    
        \multirow{2}{*}{~~~~~~~\textbf{Method}} &  \multicolumn{3}{c|}{\textbf{1-to-1 mappings}} \\\cline{2-4}
        & 
            \textbf{~~\# M~~} &  \textbf{R} & \textbf{P$\mathbf{^\approx}$}\\\hline\hline 
        LogMap &  
        $20,585$ & ~~$0.81$~~ & ~~$0.87$~~ \\\hline 
        AML &  
        $14,148$ & $0.77$ & $0.94$\\\hline 
        String similarity ($>0.8$) & 
        $20,423$ & $0.76$ & $0.87$ \\\hline \hline 
        Consensus (LogMap $\cap$ AML)  &
        $12,740$ & $0.76$ & $0.98$\\\hline
        \textbf{LogMap}  $\mathbf{\cup}$ \textbf{AML} &
        $\mathbf{21,145}$ & $\mathbf{0.83}$ & $\mathbf{0.86}$ \\\hline
    \end{tabular}
    \vspace{0.25cm}
    \caption{Alignment results for ECOTOX-NCBI. \#M: number of mappings (at instance level), R: Recall, P$^\approx$: estimated precision.}
    \label{tab:alignment_results}
\end{table}

Note that it is expected an entity from ECOTOX to match to a single entity in the NCBI Taxonomy, and vice-versa. Hence, 1-to-N 
and N-to-1 alignments were filtered according to the system computed confidence. 
A partial mapping curated by experts can be obtained through the ECOTOX Web.\footnote{ECOTOX interface: \url{https://cfpub.epa.gov/ecotox/search.cfm}} We have gathered a total of 2,321 mappings for validation purposes.
Table \ref{tab:alignment_results} shows the alignment results over the ground truth samples for 
the 
1-to-1 (filtered) system mappings. We report number of mappings (\#M), Recall (R) and estimated precision (P$^\approx$) with  respect  to  the known entities in the incomplete ground truth, assuming only 1-to-1 mappings are valid. P$^\approx$ is calculated as
\begin{align}
    P^\approx = & |M^\approx \cap M_{ref}|/|M^\approx|,  \\
    M^\approx = &  \{\langle e_e, \texttt{owl:sameAs}, e_n \rangle \in M \notag \\ 
    & \mid e_e \in \mathcal{E}_e^{ref} \lor e_n \in \mathcal{E}_n^{ref}\},
\end{align}
where $M_{ref}$ is the (incomplete) reference mapping set and $M$ is the set of generated mappings between entities $e_e\in\mathcal{E}_e$ from ECOTOX and entities $e_n\in\mathcal{E}_n$ from the NCBI Taxonomy, $\mathcal{E}_e^{ref}\subseteq\mathcal{E}_e$ and $\mathcal{E}_n^{ref}\subseteq\mathcal{E}_n$ are the sets of entities that appear in the reference mappings.
Thus, $M^\approx$ is defined as a subset of mappings from $M$ involving entities in the reference mapping set $M_{ref}$.
Recall is defined in the standard way as 
\begin{align}
    R = |M \cap M_{ref}|/|M_{ref}|.
\end{align}
Note that, the recall will be the same for $M$ and $M^\approx$.

We have selected the union of the 1-to-1 \change{equivalence\footnote{There is no need for more complex mappings in this use case.}} mappings computed by AML and LogMap to be integrated within TERA, as they represent the mapping set with the best recall with a reasonable estimated precision. This choice was made by considering the large uncertainty of downstream applications (effect prediction and risk assessment), where we prefer a larger coverage of the domain. 
See Triple \texttt{\ref{ax:et-ncbi}} in Table~\ref{tab:triples} for an example of a system computed mapping between ECOTOX and the NCBI Taxonomy.

\begin{code*}[t!]
    \begin{minted}{sparql}
PREFIX owl: <http://www.w3.org/2002/07/owl#> 
PREFIX wdt: <http://www.wikidata.org/prop/direct/> 
PREFIX wd: <http://www.wikidata.org/entity/> 
CONSTRUCT {?taxon owl:sameAs ?ncbi , ?eol .} 
WHERE {
    ?taxon wdt:P31 wd:Q16521 .
    OPTIONAL {
        ?taxon wdt:P685 ?ncbi_id .
        BIND(
            IRI(CONCAT(
               "https://www.ncbi.nlm.nih.gov/taxonomy/taxon/",
                ?ncbi_id)) 
            AS ?ncbi)
    }
    OPTIONAL {
        ?taxon wdt:P830 ?eol_id .
        BIND(IRI(CONCAT("https://eol.org/pages/",?eol_id)) AS ?eol)
    }
}
    \end{minted}
    \vspace{-0.5cm}
    \caption{Construct 
    taxon mapping between Wikidata and, NCBI and EOL. \texttt{wd:Q16521} is the class of all taxa, while \texttt{wdt:P31}, \texttt{wdt:P685} and \texttt{wdt:P830} are the relations \textit{instance of}, \textit{NCBI Taxonomy ID} and \textit{Encyclopedia of Life ID}, respectively.}
    \label{lst:ncbi2wikidata}
\end{code*}

\change{
\medskip
We use Wikidata as source of alignments between the NCBI Taxonomy and EOL, and among the used chemical datasets. Alignments are extracted via Wikidata's query interface (\ie SPARQL endpoint).\footnote{Wikidata endpoint: \url{https://query.wikidata.org/sparql}}
The data in Wikidata concerning species and chemicals are in large parts manually curated \cite{Waagmeester2020} and will have a low error rate, comparatively to using the automated ontology alignment systems. 
}

\smallskip
\noindent
\textit{Alignment between the NCBI Taxonomy and EOL}.
In order to include in TERA trait data from EOL, we need to establish an alignment between EOL and the NCBI Taxonomy. We have constructed equivalence triples between the NCBI Taxonomy and EOL identifiers using Wikidata. The species identifiers are available as literals in Wikidata. Therefore, we concatenate them with the appropriate namespace.
Listing~\ref{lst:ncbi2wikidata} represents the SPARQL CONSTRUCT query used against the Wikidata endpoint. \change{Here, we query Wikidata for instances of taxa, thereafter adding optional triple patterns for NCBI Taxonomy and EOL identifiers which are added as \texttt{owl:sameAs} triples to TERA.}

Examples of resulting mapping triples are shown in \texttt{\ref{ax:ncbi-wd}}-\texttt{\ref{ax:wd-eol}} in Table~\ref{tab:triples}. \change{The proportion of species in Wikidata where this mapping exists is $49\%$.}

\begin{code*}[t!]
    \begin{minted}{sparql}
PREFIX owl: <http://www.w3.org/2002/07/owl#> 
PREFIX wdt: <http://www.wikidata.org/prop/direct/> 
PREFIX wd: <http://www.wikidata.org/entity/> 
CONSTRUCT {?chemical owl:sameAs 
            ?cas, ?chembl, ?mesh, ?pubchem , ?inchikey .} 
WHERE {
?chemical wdt:P31 wd:Q11173 .
    OPTIONAL {
    ?chemical wdt:P231 ?cas_id .
    BIND(IRI(
        CONCAT("https://cfpub.epa.gov/ecotox/chemical/",
                REPLACE(?cas_id,'-',''))) AS ?cas)
    }
    OPTIONAL {
        ?chemical wdt:P592 ?chembl_id .
        BIND(IRI(
        CONCAT("http://rdf.ebi.ac.uk/resource/chembl/molecule/",
                ?chembl_id)) AS ?chembl)
    }
    OPTIONAL {
        ?chemical wdt:P486 ?mesh_id .
        BIND(IRI(
        CONCAT("http://id.nlm.nih.gov/mesh/",?mesh_id)) AS ?mesh)
    }
    OPTIONAL {
        ?chemical wdt:P662 ?pubchem_id .
        BIND(IRI(
        CONCAT("http://rdf.ncbi.nlm.nih.gov/pubchem/compound/CID",
                ?pubchem_id)) AS ?pubchem)
    }
    OPTIONAL {
        ?chemical wdt:P235 ?inchikey_id .
        BIND(IRI(
        CONCAT("https://rdf.ncbi.nlm.nih.gov/pubchem/inchikey/",
                ?inchikey_id)) AS ?inchikey)
    }
}
    \end{minted}
   \vspace{-0.25cm}
    \caption{Construct chemical mapping between Wikidata and ECOTOX, ChEMBL, MeSH and PubChem.
    \texttt{wdt:P31} is the predicate for \emph{instance of} and \texttt{wd:Q11173} is the class of all chemical compounds. \texttt{wdt:P231}, \texttt{wdt:P592}, \texttt{wdt:P486}, \texttt{wdt:P662} and \texttt{wdt:P235} are the relations for \textit{CAS Registry Number}, \textit{ChEMBL ID}, \textit{MeSH ID}, \textit{PubChem CID} and \texttt{InChIKey}, respectively.}
    \label{lst:cas2inchikey}
\end{code*}

\smallskip
\noindent
\textit{Alignment between chemical entities}.
The mapping between ECOTOX chemical identifiers (CAS Registry Numbers) to Wikidata entities enables the alignment to a vast set of chemical datasets, \eg PubChem, ChEBI, KEGG, ChemSpider, MeSH, UMLS, to name a few. The construction of equivalence triples between CAS, ChEMBL, MeSH, PubChem and Wikidata identifiers is shown in Listing~\ref{lst:cas2inchikey}. As for the case of species identifiers, the literal representing a chemical identifier is concatenated with the corresponding namespace. For the CAS Registry Numbers we also remove the hyphens to match ECOTOX notation. Examples of resulting mapping triples are shown in \texttt{\ref{ax:et-wd}}-\texttt{\ref{ax:wd-inchikey}} in Table~\ref{tab:triples}.

\change{
    These mappings are not complete, but for some the coverage is large. Out of the chemicals used in ECOTOX, $73\%$ have an equivalence in Wikidata (through the CAS registry numbers). Moreover, Wikidata chemicals has $4\%$ ChEMBL identifiers, $0.5\%$ MeSH identifiers, $55\%$ PubChem identifiers, and $95\%$ InChiKey identifiers.
}

\subsubsection{Taxonomy sub-KG construction}

The Taxonomy sub-KG ($KG_S$) integrates data from the NCBI Taxonomy and the EOL trait data.
The integration of the NCBI Taxonomy into the TERA knowledge graph is split into several sub-tasks.
\vspace{-0.2cm}
\begin{enumerate}[\it (i)]
        \item We load 
        the hierarchical structure included in the NCBI Taxonomy file \textit{nodes.dmp}. The columns of interest are the taxon identifiers of the child and parent taxon, along with the rank of the child taxon and the division where the taxon belongs. We use this to create triples like \texttt{\ref{ax:tax-type1}}-\texttt{\ref{ax:tax-type2}} and \texttt{\ref{ax:rank}}-\texttt{\ref{ax:sub}} in Table \ref{tab:triples}.
        \item To aid alignment between the NCBI Taxonomy and the ECOTOX identifiers, we add the synonyms found in \textit{names.dmp}. Here, the taxon identifier, its name and name type are used to create triples like \texttt{\ref{ax:name}} in Table \ref{tab:triples}. Note that a taxon in the NCBI Taxonomy can have several synonyms while a taxon in ECOTOX usually has two, \ie common name and scientific name.
        \item Finally, we add the labels of the divisions found in \textit{divisions.dmp} (see triples \texttt{\ref{ax:label}} and \linebreak \texttt{\ref{ax:label2}}). 
        We also add disjointness axioms among unrelated divisions, \eg triple \texttt{\ref{ax:disj}} in Table~\ref{tab:triples}.
    \end{enumerate}


We use the TraitBank from EOL \cite{eol2014} to add species traits to TERA. The TraitBank is modeled as a property graph and can be accessed as a \emph{neo4j} database or via a set of tabular files. To integrate the TraitBank into TERA we validate the identifiers used in EOL and convert to URIs. If an identifier is not a valid URI, we replace invalid symbols. 
%
A trait example is shown as triple \texttt{\ref{ax:habitat}} in Table~\ref{tab:triples}. The EOL TraitBank also includes 
subsumption definitions (\ie via \texttt{rdfs:subClassOf})
for a large portion of traits. These 
subsumptions
can be downloaded separately and 
are added to TERA in a similar way as mentioned~above. 

\subsubsection{Chemical sub-KG construction}



The Chemical sub-KG ($KG_C$) is created from PubChem~\cite{pubchem}, ChEMBL~\cite{ChEBI}, and MeSH~\cite{mesh}. These datasets are available for download as RDF triples. In addition, ChEMBL and MeSH can be accessed through the EBI and MeSH SPARQL endpoints, respectively. 

The chemical subset of PubChem is used since information about chemicals is standardized in PubChem, while information about substances is not. In this subset we use: 
\begin{inparaenum}[\it (i)]
\item component information, \ie what are the building blocks of the chemical or parts of a mixture;
\item type assertions, 
which either link to ChEBI or describe the type of molecule, \eg small or large;
\item role assertions, 
which describe additional attributes or relationships of the chemical, \eg \texttt{FDAApprovedDrug}; and
\item drug products, which link to the clinical data in SNOMED~CT~\cite{benson2012principles}.
%
\end{inparaenum} Examples of these can be seen in triples \texttt{\ref{ax:has_component}}, \texttt{\ref{ax:small_molecule}} and \texttt{\ref{ax:is_active}} in Table~\ref{tab:triples}.

Parent chemical data in PubChem is limited to permutations \eg bonds, polarity, and part of mixtures axioms (triple \texttt{\ref{ax:parent_compound}} in Table \ref{tab:triples}). Therefore, we use the hierarchical data about chemicals from MeSH.
In addition to this 
data, we create similarity triples between chemicals. This is impractical to download, but can be calculated on demand. We add similarity triples to TERA where the Tanimoto (Jaccard) distance between the chemical fingerprints (gathered using PubChemPy~\cite{pubchempy}) is $\geq 0.9$,\footnote{\change{Default value used in PubChem~\cite{Kim2016}.}} see triple~\ref{ax:similar_to} in Table~\ref{tab:triples}.

ChEMBL contains facts about bioactivity of chemicals. This contributes in assessing the danger of a chemical. In TERA, we use the mode of action (MoA) and target (receptor targeted by MoA; triple \texttt{\ref{ax:has_target}} in Table \ref{tab:triples}). These targets are organized in a hierarchy using \texttt{chembl:relSubsetOf} relations (see triple \texttt{\ref{ax:has_target_rel}}). 
The receptors will link to which organism it belongs to, however, we leave the inclusion of this information for future work. 

We use the entire MeSH dataset in TERA. MeSH is organised as several hierarchies. The most prominent classifications are based on chemical groups and the intended use of the chemicals. Triples \texttt{\ref{ax:mesh_hier}} and \texttt{\ref{ax:mesh_hier_func}} in Table \ref{tab:triples} show examples of chemical group and functional classifications.

\begin{code*}[t!]
    \begin{minted}{sparql}
PREFIX rdfs:  <http://www.w3.org/2000/01/rdf-schema#> .
PREFIX rdf:  <http://www.w3.org/1999/02/22-rdf-syntax-ns#> .
PREFIX eol: <http://eol.org/schema/terms/> .
PREFIX et:  <https://cfpub.epa.gov/ecotox/> .
PREFIX et_endpoint:  <https://cfpub.epa.gov/ecotox/endpoint/> .
PREFIX et_effect:  <https://cfpub.epa.gov/ecotox/effect/> .
PREFIX qudt:  <http://qudt.org/schema/qudt#> .
SELECT ?s ?c ?conc ?concunit 
WHERE {
    ?s  eol:endemicTo [ rdfs:label "Oslofjorden"@no ] .
    _:b a et:Test ;
        et:species ?s .
        et:chemical ?c .
        et:hasResult [
            et:endpoint et_endpoint:LC50 ;
            et:effect et_effect:Mortality ;
            et:concentration [ 
                               rdf:value ?conc ;
                               qudt:units ?concunit 
                             ] .
        ]
} 
    \end{minted}
    \vspace{-0.5cm}
    \caption{Query to select all species, chemicals, concentrations and units, where the species is endemic to the \emph{Oslofjord}.}
    \label{lst:example}
\end{code*}

\subsection{TERA for data access}
\label{sec:dataaccess}

 TERA covers knowledge and data 
 relevant to
 the ecotoxicological domain and enables an integrated 
 semantic access across data sets. In addition, the adoption of 
 an RDF-based knowledge graph enables the use of an extensive range of Semantic Web infrastructure (\eg reasoning engines, ontology alignment systems, SPARQL query engines).
 
 The data integration efforts and the construction of TERA go 
 in line with the vision in the computational risk assessment communities (\eg Norwegian Institute for Water Research's Computational Toxicology Program (NCTP)), where increasing the availability and accessibility of knowledge enables optimal decision making. 


The knowledge in TERA can be accessed via predefined queries\footnote{Predefined queries are typically abstractions of SPARQL queries.} (\eg classification, sibling, and name queries, and fuzzy queries over the species names) and arbitrary SPARQL queries. 
The (final) output is flexible to the
task, and can be given either as a graph or in tabular format. Listing \ref{lst:example} shows an example query to extract the chemicals and concentrations, at which, the species in the \emph{Oslofjord} experience lethal effects.


\subsection{TERA for effect prediction} 
\label{sec:teraprediction}

TERA is used as background knowledge in combination with machine learning models for chemical effect prediction. 
TERA's sub-KGs play different roles in effect prediction. 
The rich semantics of the species and chemical entities in the Taxonomy sub-KG ($KG_S$) and the Chemical sub-KG ($KG_C$), respectively, are 
embedded into low-dimensional vectors;
while the Effects sub-KG ($KG_E$) provides the 
training samples for the prediction model. 
Each sample is composed of a chemical, a species, a chemical concentration, and the outcome or endpoint of the experiment. 
More details are given in Section \ref{sec:prediction}, where the effect prediction model is built upon state-of-the-art knowledge graph embedding models.


\setlength{\tabcolsep}{4.1pt}
\begin{table}[t!]
    \centering
    \begin{tabular}{|l|c|c|c|c|c|}
        \hline 
        \multicolumn{1}{|c|}{\textbf{Dataset}} & ~~$\mathbf{RD}$~~ & ~~$\mathbf{ED}$~~ & ~~$\mathbf{RE}$~~ & ~~$\mathbf{EE}$~~ & ~~$\mathbf{AD}$~~ \\\hline\hline 
        TERA $\text{KG}_C$\hspace{5pt} & $2.3\times 10^5$ & $5.5$ & $3.0$ & $24$ & $4.6\times 10^{-7}$\\
        TERA $\text{KG}_S$ & $6.6\times 10^4$ & $5.1$ & $2.7$ & $23$ & $3.7\times 10^{-7}$ \\ \hline\hline  
        TERA $\text{KG}_C^\prime$\hspace{5pt} & $6.9\times 10^3$ & $8.6$ & $2.3$ & $17$ & $7.7\times 10^{-5}$\\
        TERA $\text{KG}_S^\prime$ & $3.8\times 10^2$ & $15$ & $2.3$ & $14$ & $8.9\times 10^{-4}$ \\ \hline\hline  
        YAGO3-10 & $2.9\times 10^{4}$ & $18$ & $2.0$ & $20$ & $7.1\times 10^{-5}$\\
        FB15k-237 & $1.3\times 10^3$ & $43$ & $4.5$ & $16$ & $1.3\times 10^{-3}$\\
        WN18 & $8.4\times 10^3$ & $7.4$ & $2.1$ & $16$ & $9.0\times 10^{-5}$\\
        WN18RR & $8.5\times 10^3$ & $4.5$ & $1.5$ & $19$ & $5.5\times 10^{-5}$\\\hline
    \end{tabular}
    \medskip
    \caption{Densities and entropies of benchmark datasets. TERA $KG_C$ and $KG_S$ are the chemical and species parts of TERA, while $KG^{\prime}_C$ and $KG^{\prime}_S$ denote the parts of TERA used in prediction in Section \ref{sec:results}.}
    \label{tab:stats}
\end{table}

 Table \ref{tab:stats} 
 shows the sparsity-related measures of common benchmark datasets\footnote{YAGO3-10 \cite{yago}, FB15k-237 \cite{10.1145/1376616.1376746}, WN18 \cite{10.1145/219717.219748} and WN18RR~\cite{dettmers2018}.} and TERA's $KG_C$ and $KG_S$ (triples involving literals are removed).
We follow Pujara et al. \cite{pujara-etal-2017-sparsity} and calculate the relational density, ${RD}=|\mathcal{T}|/|\mathcal{R}|$, and entity density, ${ED}=2|\mathcal{T}|/|\mathcal{E}|$, where $\mathcal{T}$, $\mathcal{R}$, and $\mathcal{E}$ are the sets
of triples, relations, and entities in 
the knowledge graph, respectively.
The entity entropy (EE) and the relation entropy (RE) indicate whether there are biases 
(the lower EE or RE, the larger bias)
in the triples in the KG \cite{pujara-etal-2017-sparsity}, and 
are calculated as 
\begin{align}
    P(r) &= \frac{|t.p = r|}{|T|},\\
    P(e) &= \frac{|t.sb=e|+|t.ob=e|}{|T|}, \\
    {RE} &= \sum_{r\in \mathcal{R}}-P(r)log(P(r)),\\
    {EE} &= \sum_{e\in \mathcal{E}}-P(e)log(P(e)),
\end{align}
where $|t.p=r|$ is the number of triples with $r$ as predicate, and $|t.sb=e|+|t.ob=e|$ is the number triples with $e$ as subject or object.

In addition, we calculate the absolute density of the graph, which is~${AD}=|\mathcal{T}|/(|\mathcal{E}|(|\mathcal{E}|-1))$. This is the ratio of edges to the maximum number of edges possible in a simple directed graph \cite{doi:10.1137/0720013}.

High RD and low RE typically lead to a worse performance, while high ED and low EE often lead to better link prediction performance (\eg \cite{dettmers2018}).
%
In Table \ref{tab:stats} we can see that the density and entropy values are in between those for YAGO3-10 and FB15k-237, which typically lead to worse and better predictive performance, respectively~\cite{dettmers2018}.
This shows that TERA is a suitable background knowledge to extrapolate effect data and, at the same time, 
an interesting dataset to benchmark state-of-the-art knowledge graph embedding models. Note that using the full TERA (\ie $KG_C$ and $KG_S$), according to RD, will be more challenging than using the reduced TERA fragments (\ie $KG^{\prime}_C$ and $KG^{\prime}_S$) for prediction.
Full details of the construction of $KG^{\prime}_C$ and $KG^{\prime}_S$ are given in Section \ref{sec:dataprep}.

\section{Adverse biological effect prediction}
\label{sec:prediction}

The aim of chemical effect prediction is to extrapolate exiting data to new   combinations of (possibly unknown) chemicals and species. In this section we present three classification models used to predict the adverse biological effect of chemicals on species: 
\begin{inparaenum}[\it (i)]
\item a multilayer perceptron (MLP) model (our baseline),
\item the baseline model fed with pre-trained KG embeddings,
\item a model that simultaneously trains the baseline model and the KGE models (\ie it fine-tunes the KG embeddings).
\end{inparaenum}
\change{A MLP was chosen as baseline as it is a basic model where additional components and penalties can be easily added and assessed as we do in our third model (see Section \ref{sec:fine-tuned-model}).}

The models have three inputs, namely a chemical $c$, a species $s$, and a chemical concentration $\kappa$ (denoted $x_{c,s,\kappa}$). 
The output is a binary value that represents whether the chemical at the given concentration has a lethal effect on the species:
%
\begin{align}
    y_{c,s,\kappa} = 
    \begin{cases}
    1 & c \text{ is lethal to } s \text{ at } \kappa, \\
    0 & \text{otherwise.}
    \end{cases} \label{eq:effect_input}
\end{align}
Note that the effect can have a more fine-grained categorization (endpoints LC$x$, LD$x$, EC$x$\footnote{If effect is mortality (\eg see Table~\ref{tab:ecotox_ex2}).}, and NR-LETH in Table \ref{tab:endpoints}). Without 
losing the generality in introducing and evaluating our effect prediction methods, we simplify the effect into two cases: ``lethal'' and ``non-lethal''.  




\medskip
\noindent
\textit{Notation.}
Throughout this section we use bold lower case letters to denote vectors while matrices are denoted as bold upper case letters. 
The vector representation of an entity and a relation are noted as $\vec{e}_e$ and $\vec{e}_p$, respectively. These vectors are either in $\mathbb{R}^k$ or $\mathbb{C}^k$, where $k$ is the embedding dimension.

\begin{figure*}[t]
\begin{subfigure}{0.44\textwidth}
\centering
    \centering
    \includegraphics[width=0.99\textwidth]{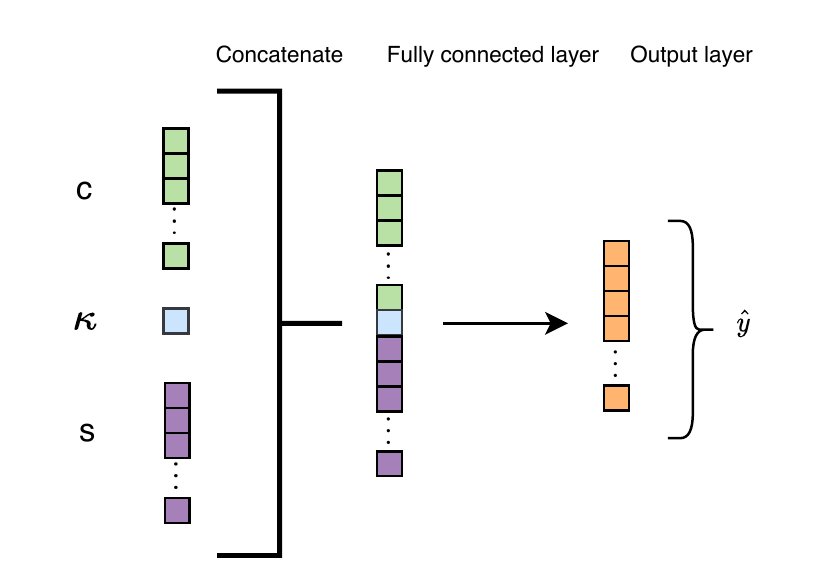}
    \caption{\footnotesize{Simple setting. Without transformation layers: $n_c=0, n_s=0, n_{\kappa}=0$ and $n=1$.}}
    \label{fig:model_no_hidden}
\end{subfigure}
~~~
\begin{subfigure}{0.52\textwidth}
    \centering
    \includegraphics[width=0.99\textwidth]{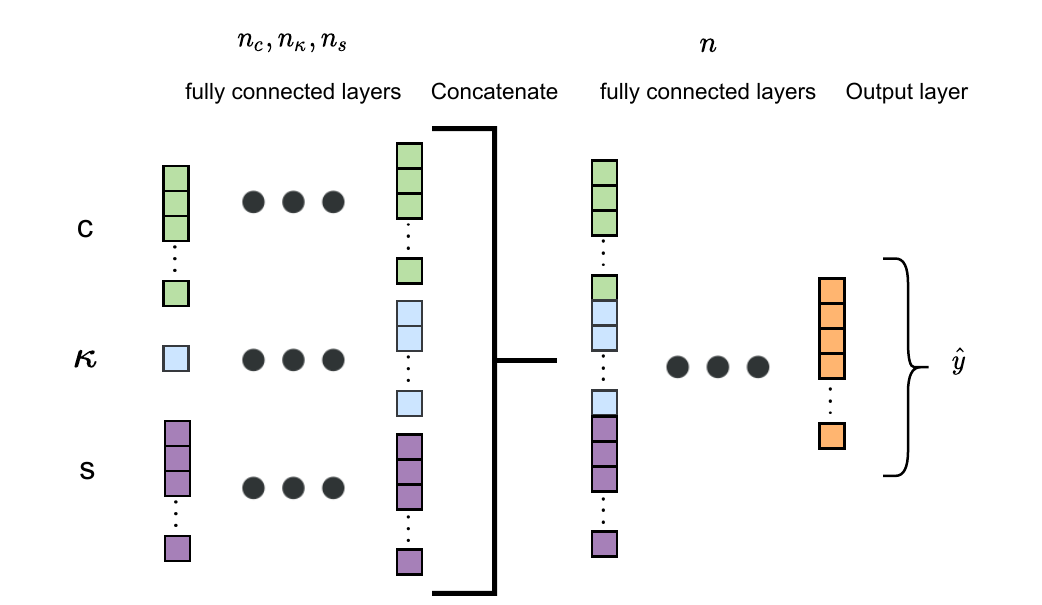}
    \caption{\footnotesize{Complex setting. Model with branches/transformation layers. In contrast to the simple setting, here $n_c\geq 1, n_s\geq 1, n_{\kappa}\geq 1$ and $n\geq 1$. }}
    \label{fig:model_with_hidden}
\end{subfigure}
\caption{Baseline model. Inputs: $c,s,\kappa$  as in Equation \eqref{eq:mlp1}; Outputs: $\hat{y}$ as in Equation \eqref{eq:mlp6}. }\label{fig:baseline}
\end{figure*}

\subsection{Baseline model}
\label{sec:baseline}
Our baseline prediction model is a multilayer perceptron (MLP) with multiple hidden layers. 
$n_c$ hidden layers are appended to the embedding $\vec{e}_c$ of the chemical $c$, $n_s$ hidden layers are appended to the embedding $\vec{e}_s$ of species $s$, and $n_{\kappa}$ hidden layers appended to the real valued chemical concentration $\kappa$. Thereafter, $n$ hidden layers are further appended to the output of the previous hidden layers concatenated. 
Specifically, the model can be expressed by the following equations (with 
$x_{c,s,\kappa}$
as input):

\begin{align}
    \vec{y}^0_c &= \vec{e}_c,\ \vec{y}^0_s = \vec{e}_s,\ y_\kappa^0 =\kappa \label{eq:mlp1} \\ 
    \vec{y}^h_c &= ReLu(\vec{y}^{h-1}_c \mat{W}^h_c + \vec{b}^h_c),\ h \in \{0,\dots,n_c\}\label{eq:mlp2} \\
    \vec{y}^h_s &= ReLu(\vec{y}^{h-1}_s \mat{W}^h_s + \vec{b}^h_s),\ h \in \{0,\dots,n_s\}\label{eq:mlp3} \\
    \vec{y}^h_\kappa &= ReLu(\vec{y}^{h-1}_\kappa \mat{W}^h_\kappa + \vec{b}^h_
    \kappa),\ h \in \{0,\dots,n_\kappa\}\label{eq:mlpkappa} \\
    \vec{y}^0 &= \left[\vec{y}^{n_c}_c,\vec{y}^{n_s}_s,\vec{y}^{n_\kappa}_\kappa\right]\label{eq:mlp4} \\
    \vec{y}^h &= ReLu(\vec{y}^{h-1} \mat{W}^h + \vec{b}^h),\ h \in \{1,\dots,n\}\label{eq:mlp5} \\
    \hat{y} &= \sigma(\vec{y}^n \mat{W}^n + \vec{b}^n)\label{eq:mlp6}
\end{align}
$\vec{e}_c,\vec{e}_s \in \mathbb{R}^k$ in \eqref{eq:mlp1} denote the embeddings of $c$ and $s$ respectively, and are calculated as
\begin{align}
    \vec{e}_c = \vec{\delta}_c \mat{W}_c,~~ \vec{e}_s = \vec{\delta}_s \mat{W}_s \label{eq:input_layer}
\end{align}
where $\vec{\delta}_c$ and $\vec{\delta}_s$ denote the one-hot encoding vectors of the chemical entity $c$ (w.r.t. all the entities in $\mathcal{E}_C$ from $KG_C$) and the species entity $s$ (w.r.t. all the  entities in $\mathcal{E}_S$ from $KG_S$), respectively;\footnote{$\vec{\delta}_c\in \mathbb{R}^{|\mathcal{E}_C|}$, where $\vec{\delta}_c^i=1$ if $c$ is the  $i^{th}$ chemical in  $\mathcal{E}_C$, else $0$. $\vec{\delta}_s$ is defined similarly.} $\mat{W}_c \in \mathbb{R}^{|\mathcal{E}_C|\times k}$ and $\mat{W}_s \in \mathbb{R}^{|\mathcal{E}_S|\times k}$ are embedding transformation matrices to learn.
\eqref{eq:mlp2}, \eqref{eq:mlp3} and \eqref{eq:mlp5} represent the hidden layers, where $ReLu$ denotes the rectifier function (\ie $ReLu(x)=\max(0,x)$), $\mat{W}^t_c$, $\mat{W}^t_s$ and $\mat{W}^t$ denote the weights, $\vec{b}^t_c$, $\vec{b}^t_s$ and $\vec{b}^t$ denote the biases.
$\left[\cdot, \cdot \right]$ in \eqref{eq:mlp4} denotes vector concatenation.
$\sigma$ in \eqref{eq:mlp6} denotes the sigmoid function (\ie $\sigma(x)=1/(1+\exp(-x))$).
Note that a dropout and a normalization layer is stacked after each hidden layer for regularization.

We differentiate between two settings of the baseline model (see Figure \ref{fig:baseline}):
\begin{enumerate}[\it (i)]
 \item \textit{Simple setting.} Figure \ref{fig:model_no_hidden} shows the model without embedding transformation layers, \ie $n_s=n_c=n_{\kappa}=0$, and $n=1$.
 \item \textit{Complex setting.} The complex model shown in Figure \ref{fig:model_with_hidden} introduces transformation layers on the embeddings and chemical concentration input. These transformations aim at extracting the important information in the inputs and disregard the redundant information based on the output. 
\end{enumerate} 

In the experiments we refer to the baseline models as \emph{Simple one-hot} and \emph{Complex one-hot}, depending on the selected MLP setting. 



\subsection{Baseline model with pre-trained KG embeddings}

This models relies on pre-trained embeddings of chemicals and species computed using state-of-the-art KGE models (see Section \ref{sec:relatedKGEM} and Appendix \ref{sec:appendix_kge} for an overview). 
A (different) KGE model is applied to the chemicals $KG_C$ and the species $KG_S$.

These pre-trained KG embeddings are then given as input instead of the one-hot encoding vectors in the baseline model. 
We replace the trainable matrices $\mat{W}_c$ and $\mat{W}_s$ in Equation \eqref{eq:input_layer} by the matrices composed of embeddings by the respective KGE models. 
Namely $\mat{W}_c$ is set to $[\vec{e}_{c,1};\vec{e}_{c,2};\dots;\vec{e}_{c,|\mathcal{E}_C|}]$, $\mat{W}_s$ is set to $[\vec{e}_{s,1};\vec{e}_{s,2};\dots;\vec{e}_{s,|\mathcal{E}_S|}]$, where $[\cdot;\cdot]$ denotes stacking vectors, $\vec{e}_{c,i}$ denotes the embedding of the $i^{th}$ chemical in the chemicals $KG_C$, $\vec{e}_{s,i}$ denotes the embedding of the $i^{th}$ species in the species $KG_S$.

In the experiments we refer to these models as \emph{Simple PT KGE$_C$-KGE$_S$} and \emph{Complex PT KGE$_C$-KGE$_S$}, depending on the selected MLP setting, where PT stands for pre-trained, and KGE$_C$ and KGE$_S$ are the KGE models used for the chemicals KG and the species KG, respectively (\eg \emph{Complex PT DistMult-HAKE}). For simplicity, we also refer to these models as PT-based models. 


\begin{figure*}
    \centering
    \includegraphics[width=0.95\textwidth]{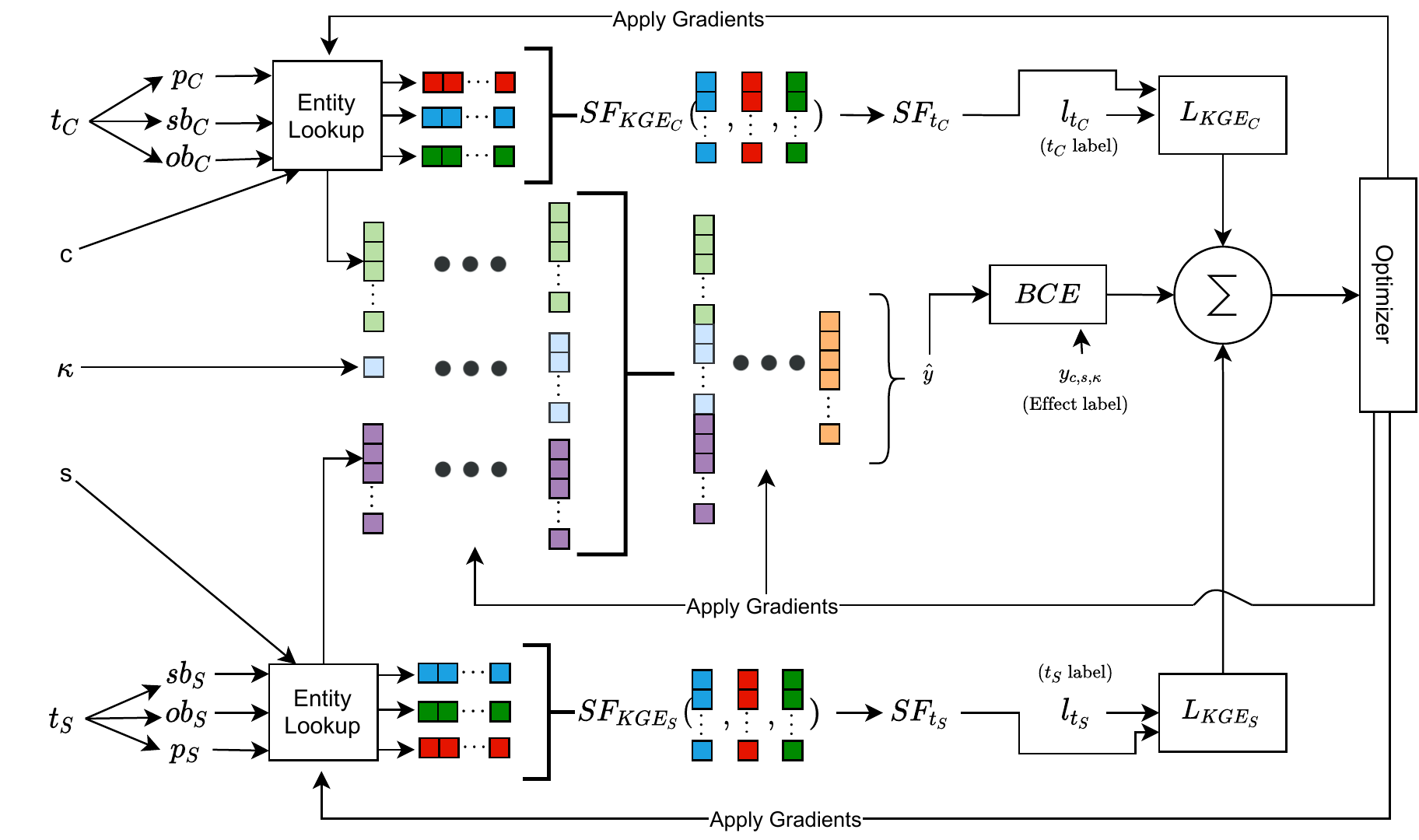}
    \caption{Fine-tuning optimization model. In addition to variables described in Figures \ref{fig:model_no_hidden} and \ref{fig:model_with_hidden}, $t_C = (sb_C,p_C,ob_C) \in KG_C \cup \overline{KG}_C$, $t_S = (sb_S,p_S,ob_S) \in KG_S \cup \overline{KG}_S$. Entity lookups transform an entity into a vector (see Equation \eqref{eq:input_layer}).
    $SF_{KGE_C}$ and $SF_{KGE_S}$ are the triple scoring functions implemented by the selected KGE model (see Appendix \ref{sec:appendix_kge}). $SF_{t_C}$ and $SF_{t_S}$ are the scores for a chemicals and species triple, respectively. 
    $x_{c,s,\kappa}$
    is the prediction input and $y_{c,s,\kappa}$ is described in Equation \eqref{eq:effect_input}.
    $l_{t_C}$ and $l_{t_S}$ are the triple labels (\ie True or False).
    $BCE$ is the binary cross-entropy loss function (from Equation \eqref{eq:bce}). The summation of the losses is described in Equation \eqref{eq:weighted_loss}, that is the loss used by the optimizer to apply changes to model weights. 
    }
    \label{fig:model_finetune}
\end{figure*}

\subsection{Fine-tuning optimization model}
\label{sec:fine-tuned-model}

This model improves upon the pre-trained KG embeddings with fine-tuning based on the effect prediction data. This is done by simultaneously training the (selected) KGE models and the MLP-based baseline model. Such that the $\mat{W}_C$ and $\mat{W}_S$, and the MLP weights ($\mat{W}_x$ and $\vec{b}_x$ in Equations (\ref{eq:mlp2}), (\ref{eq:mlp3}), (\ref{eq:mlp5}) and (\ref{eq:mlp6})) are optimized simultaneously. Note that we initialize the KGE models with the previously pre-trained embeddings.

%
The model architecture is shown in Figure \ref{fig:model_finetune} and the overall loss to minimize is
\begin{align}
    L = \alpha_{C}L_{KGE_C} + \alpha_{S}L_{KGE_S} + \alpha_{MLP} L_{MLP} \label{eq:weighted_loss} 
\end{align}
where $L_{KGE_C}$ and $L_{KGE_S}$ respectively denote the loss of the chemical KG$_C$ and the species KG$_S$ when a specific KGE model is used,\footnote{Appendix \ref{sec:loss_functions} introduces the used loss-functions in this work. The selection of the loss function for a KGE model will be via a hyper-parameter.} $\alpha_{C}$ and $\alpha_{S}$ denote their weights respectively,
$L_{MLP}$ and $\alpha_{MLP}$ denote the loss of the MLP and its weight.
Specifically, we use binary cross-entropy (BCE) 
as the loss for the classification. $L_{MLP}$ is calculated as
%
\begin{align}
\hspace{-0.35cm}
L_{MLP} = -\frac{1}{N}\sum_i^N y_i \log(\hat{y}_i) + (1-y_i)\log(1-\hat{y}_i) \label{eq:bce}
\end{align}
where $N$ 
denotes the size of training samples, $y_i$ and $\hat{y}_i$ denote the sample label and the MLP output, respectively (as in Equation \eqref{eq:effect_input}).
With the overall loss, gradient-based learning algorithms such as Adam optimizer \cite{Adam_article} can be adopted to jointly training the embeddings of both KGEs and the MLP.

Figure \ref{fig:model_finetune} 
shows the full simultaneous fine-tuning model and the optimization process. The initial state of the entity lookups is the pre-trained embeddings. 
The full training procedure is summarised as follows:
\begin{enumerate}
    \item Select $N$ triples from $KG_C$ and $KG_S$, where $N$ is the length of the effects training set.\footnote{Section \ref{sec:exp-setup} describes how the known effect data extracted from ECOTOX is split into training, validation and test sets.} 
    \item Generate negative knowledge graph triples (see Appendix \ref{sec:loss_functions} for details) from the extracted subsets of triples from $KG_C$ and $KG_S$, these negative KGs triples are referred to as $\overline{KG}_C$ and $\overline{KG}_S$. 
    \item Feed-forward the input through the model and calculate loss for each model component and combine according the loss weights. 
    \item Optimize the KG entity and relation embeddings, and the MLP layers. 
\end{enumerate}
These steps are repeated until the loss (only $L_{MLP}$) over the validation set stops improving.  

In the experiments we refer to these models as \emph{Simple FT KGE$_C$-KGE$_S$} and \emph{Complex FT KGE$_C$-KGE$_S$}, depending on the selected MLP setting, where FT stands for fine-tuning, and KGE$_C$ and KGE$_S$ are the KGE models used for the chemicals KG and the species KG, respectively (\eg \emph{Simple FT HAKE-HAKE}). For simplicity, we also refer to these models as FT-based models.

\section{Results}
\label{sec:results}
\subsection{Experimental setup}
\label{sec:exp-setup}
All models are implemented using Keras \cite{chollet2015keras} and the model codes are available in our GitHub repository, alongside all data preparation and analysis scripts.\footnote{\url{https://github.com/NIVA-Knowledge-Graph/KGs_and_Effect_Prediction_2020}} 

\subsubsection{Preparation of TERA for prediction}
\label{sec:dataprep}

As shown earlier, TERA consists of three sub-KGs. These are the basis for the chemical effect prediction.\footnote{All data used to create TERA was downloaded on the 14th of May 2020.}
We process the sub-KGs further to limit their size by  removing irrelevant triples for prediction. This is necessary to scale up the training of the KGE models. The reduction of TERA's sub-KGs is performed according to the following steps:
\begin{enumerate}[\it (i)]
    \item Effect data. For prediction purposes, the effect data in $KG_E$ is limited to four features, namely, chemical, species, chemical concentration, and effect. 
    The chemical concentrations ($\kappa$, converted to $mg/L$) are log-normalized to remove the large discrepancy in scales. As mentioned, we separate the effects into two categories for simplicity, lethal and non-lethal effects. This reduces the possibility of ambiguity among the effects that does not cause death in the test species. We label lethal effects as $1$ and non-lethal effects as $0$
    \item $KG_C$. For each chemical in the effect data, we extract all triples connected to them using a directed crawl. 
    This reduces the size of $KG_C$ to a manageable size for the KGE models. Moreover, we do not deem triples not directly connected to the effect data relevant for the prediction task, and may introduce 
    unnecessary
    noise. As mentioned before, PubChem contains similarities between chemicals based on chemical fingerprints, however, for our use-case it is unpractical to query them from the PubChem RDF data, therefore, we calculate similarity triples based on queried PubChem fingerprints. We use the same similarity threshold as PubChem, 
    \ie~$0.9$~\cite{Kim2016}.
    \item $KG_S$. The same steps as for $KG_C$ 
    are conducted for all species in the effect data. 
\end{enumerate}
\change{A simple directed crawl over all predicates is sufficient to gather the interesting data in this setting as both $KG_C$ and $KG_S$ are primarily hierarchical and we start the crawls at the leaf nodes.} 

These steps reduce $KG_C$ to $241,442$ triples and $KG_S$ to $59,673$ triples. Some statistics of $KG_C$ and $KG_S$, and the reduced fragments $KG_C^{\prime}$ and $KG_S^{\prime}$, are given in Table \ref{tab:stats} (Section \ref{sec:teraprediction}). In the rest of the paper were refer to TERA's reduced sub-KGs simply as $KG_C$ and $KG_S$.

\change{
The transformation from 
TERA's $KG_C$ and $KG_S$
to model input is done by first dropping literals, thereafter assigning each entity 
an unique integer identifier which corresponds to the index of a column vector in matrices $\mat{W}_c$ or $\mat{W}_s$ in Equation \eqref{eq:input_layer}, depending on which sub-KG is transformed.\footnote{$i \in [0,|\mathcal{E}_C|-1]$ for $KG_C$ and $i \in [0,|\mathcal{E}_S|-1]$ for $KG_S$} 
Relations are treated similarly. 
}

\subsubsection{Sampling}
\label{sec:sampling-strategies}
We use four sampling strategies
of the effect data to analyze how the proposed classification models behave by varying the data parts that are used for training and testing. \change{Note that, we only consider effect data where the chemical and species have mappings to external sources (\eg NCBI Taxonomy and Wikidata, \cf Section \ref{sec:alignment}) so that there is additional contextual information that can be used by the KGE models.}
For each of the strategies, the validation and test sets contain unseen chemical-organism pairs with respect to the training set. The strategies, however, differ with respect to the individual organism and chemical as follows:
\begin{enumerate}[\text{Strategy} \it (i)]
    \item Random $70\%/15\%/15\%$ training/validation/test split on the entire dataset (\ie the chemicals and the organisms in the validation and test will most probably be known). 
    \item Training/validation/test split where there is no overlap between chemicals in the three sets (\ie the chemicals in the validation and test sets are unknown). This resulted on a $77\%/14\%/9\%$~split.
    \item Training/validation/test split where there is no overlap between species in the three sets (\ie the species in the validation and test sets are unknown).
    This resulted on a $77\%/14\%/9\%$ split.
    \item Training/validation/test split with no chemicals or species overlap in the three sets (\ie both the chemicals and the organisms in the validation and test sets are unknown). This resulted on a $72\%/14\%/14\%$ split.
\end{enumerate}
\change{
Note that since we use the species and chemicals as groups to divide the data rather than the samples, the splits can vary.
}
For strategies \textit{(i-iii)} there is a total of 14,377 effect data samples while for strategy \textit{(iv)} the total number samples is 5,621. 
As above, this discrepancy is down to the way we split the data. We do not split across samples, but across chemicals and species. For example, some chemicals are used on (close to) all species, therefore, these chemicals are 
discarded in the sampling strategy \textit{(iv)}, affecting the final number of samples.

There were originally 57,560 samples, however, this includes experiment duplicates, \ie same chemical, species, and endpoint, with different chemical concentrations. 
This is down to large discrepancies in laboratory testing variance, therefore, we use the median concentration across the duplicates. The prior probability is approximately $0.16/0.84$ (\ie $\approx 16\%$ of samples are labelled as non-lethal and $\approx 84\%$ of samples are labelled as lethal) across all sampling methods. We solve this when training by randomly oversampling the minority class until the prior probabilities are $0.5/0.5$ in the training set. In this case, the oversampling is performed by adding duplicates samples labelled as non-lethal. \change{Oversampling is a well established technique used in many classification problems to remove bias during learning \cite{DBLP:journals/corr/BrancoTR15}.}

\begin{table}[b]
    \centering
    \begin{tabular}{|l|r|}
        \hline
        \textbf{KGE hyper-parameters} & \textbf{Search space} \\\hline
        Loss function & $\{L_{H_1}, L_{H_2}, L_{L_1} ,L_{L_2}\}$ \\\hline
        Margin (only hinge loss) & $\{1,2,\dots,10\}$\\\hline
        Bias (only geometric models) & $\{0,1,\dots,20\}$\\\hline 
        Embedding dimension & $\{100,101,\dots,400\}$\\\hline
        Negative samples & $\{10,11,\dots,100\}$ \\\hline\hline
        \textbf{Prediction hyper-parameters} & \textbf{Search space}\\\hline 
        $n_c$ \eqref{eq:mlp2}, $n_s$ \eqref{eq:mlp3}, $n_{\kappa}$ \eqref{eq:mlpkappa}, $n$ \eqref{eq:mlp5} & $\{0,1,2,3\}$ \\\hline
        $\#$ units \eqref{eq:mlp2}, \eqref{eq:mlp3}, \eqref{eq:mlp5} & $\{2^u\ \text{with}\ u\in \{4,5,\dots,10\}\}$ \\\hline 
        $\#$ units \eqref{eq:mlpkappa} & $\{2^u\ \text{with}\ u\in \{2,3,4,5\}\}$ \\\hline
    \end{tabular}
    \vspace{0.2cm}
    \caption{Hyper-parameter choices for the models. Please refer to the Equations \eqref{eq:mlp1}-\eqref{eq:mlp6} in Section \ref{sec:baseline} 
    for the prediction hyper-parameters.}
    \label{tab:hp}
\end{table}

\begin{table*}[t]
    \centering
    \begin{tabular}{|l|c|c|c|c|c|}
        \hline
        \textbf{Model} & \textbf{Loss function} & \textbf{Margin} & \textbf{Bias} & \textbf{Embedding dimension} & \textbf{Negative Samples} \\\hline 
        DistMult & $L_{L_2}$~/~$L_{H_2}$ & -~/~$2$ & - & $143$~/~$383$ & $28$~/~$43$ \\\hline 
        ComplEx & $L_{L_2}$~/~$L_{H_2}$ & -~/~$4$ & - & $163$~/~$372$ & $27$~/~$42$ \\\hline 
        HolE & $L_{H_2}$~/~$L_{L_2}$ & $6$~/~- & -  & $188$~/~$376$ & $30$~/~$100$ \\\hline 
        TransE & $L_{H_2}$~/~$L_{H_1}$ & $4$~/~$7$ & $14$~/~$20$ & $226$~/~$196$ & $23$~/~$57$ \\\hline
        RotatE & $L_{H_2}$~/~$L_{H_2}$ & $5$~/~$2$ & $16$~/~$6$ & $271$~/~$398$ & $75$~/~$22$ \\\hline
        pRotatE & $L_{L_2}$~/~$L_{L_2}$ & -~/~- & $14$~/~$16$ & $164$~/~$210$ & $34$~/~$82$ \\\hline 
        HAKE & $L_{L_2}$~/~$L_{L_2}$ & -~/~- & $12$~/~$10$ & $108$~/~$359$ & $56$~/~$13$ \\\hline
        ConvKB & $L_{L_2}$~/~$L_{H_2}$ & -~/~$5$ & - & $248$~/~$276$ & $18$~/~$90$ \\\hline  
        ConvE & $L_{H_1}$~/~$L_{H_1}$ & $7$~/~$3$ & - & $228$~/~$196$ & $68$~/~$40$ \\\hline 
    \end{tabular}
    \vspace{0.2cm}
    \caption{Best hyper-parameters for KGE models. The two values before and after ~/~ are for the embeddings
    of $KG_C$ and $KG_S$, respectively.}
    \label{tab:hp_best}
\end{table*}

\subsubsection{Hyper-parameters}
To optimize the hyper-parameters for the KGE and classification models we use random search over the parameter ranges. We conduct 20 trials per model. Tables \ref{tab:hp} and \ref{tab:hp_best} contain the best hyper-parameters and can be used to reproduce the top performing models.

To find the best hyper-parameters for the KGE models, we use the loss as a proxy for performance, normalized by the initial loss, $RL_{ep} = L_{ep}/L_0$, where $L_{ep}$ is the training loss at epoch $ep$, $L_0$ is the loss with the initial weights.

We use validation loss to select the best hyper-parameter setting for the classification models presented in Section~\ref{sec:prediction}.
The best prediction models are refitted and evaluated 10 times to reduce the influence of initial conditions on the metrics. The average and standard deviation of the metrics are presented in Section~\ref{sec:prediction_results}. 

The hyper-parameter ranges for the KGE models are shown in Table \ref{tab:hp} based on common values used in the literature. We conduct 20 trials of random hyper-parameters choices and validate over the validation data.  In Table \ref{tab:hp_best} we show the best hyper-parameters.

\begin{table*}[t]
    \centering
    \begin{tabular}{|l|c|c|}
    \hline 
    \textbf{Model} & \textbf{Sampling} & \textbf{\# units} \\\hline 
    \multirow{4}{*}{Complex one-hot} 
    & \textit{(i)} & $(128)/(128)/-/-$\\ 
    & \textit{(ii)} & $(128)/(256)/(8,8)/-$ \\ 
    & \textit{(iii)} & $(256,128)/(128)/(4,4,4)/-$ \\ & \textit{(iv)} & $(256,256)/(128)/(8,8)/(128)$ \\\hline 
    Complex PT DistMult-HAKE (top-1 in \textit{(i)}) & \textit{(i)} & $(256,256)/(256)/(16,4)/(512,64)$ \\\hline 
    Complex PT HolE-ConvKB (top-1 in \textit{(ii)}) &\textit{(ii)} & $(512,128,128)/(512)/-/(64)$ \\\hline 
    \multirow{2}{*}{Complex PT HAKE-DistMult (top-1 in \textit{(iii,iv)})}  
    & \textit{(iii)} & $(64)/(512)/(16,32)/(16)$\\
    & \textit{(iv)} & $(128)/-/(4,8,8)/(256,128)$ \\\hline 
    \end{tabular}
    \vspace{0.2cm}
    \caption{Number of units in the hidden layers in the (complex) one-hot model and the top-1 prediction models with pre-trained KG embeddings. The same parameters are used for the fine-tuning models. 
    Organized as follows: $(|\vec{b}_c^1|,...,|\vec{b}_c^{n_c}|)/(|\vec{b}_s^1|,...,|\vec{b}_s^{n_s}|)/(|\vec{b}_{\kappa}^1|,...,|\vec{b}_{\kappa}^{n_\kappa}|)/(|\vec{b}^1|,...,|\vec{b}^n|)$ as in Equations (\ref{eq:mlp2}), (\ref{eq:mlp3}), (\ref{eq:mlpkappa}), and (\ref{eq:mlp5})). $-$ denotes no hidden layers. \eg $(128)/(256)/(8,8)/-$ denotes $n_c=1,n_s=1,n_{\kappa}=2,n=0$ and $|\vec{b}|_c^1=128$, $|\vec{b}_s^1|=256$,   $|\vec{b}_{\kappa}^1|=8$ and $|\vec{b}_{\kappa}^2|=8$. 
    }
    \label{tab:pred_hp}
\end{table*}


We can see in Table \ref{tab:hp_best} that the decomposition models have similar hyper-parameters for $KG_C$ and $KG_S$. As shown in Section~\ref{sec:teraprediction}, the major difference between $KG_C$ and $KG_S$ is the relational density. Therefore, 
it is reasonable to believe that a lower relational density KG requires more parameters to have an equivalent representation in the embedding space.
We can 
get the same observation for the geometric models except for TransE, where the embedding dimensions are similar.
ConvE is more efficient in embedding dimension than ConvKB, however, since ConvE is slightly more complex than ConvKB this is expected. 
The difference in negative samples could be down to our implementation of ConvE, which varies from the original. Our implementation of all models 
relies on 1-to-1 scoring of triples, while the implementation of ConvE originally used 1-to-$|\mathcal{E}|$ scoring, where $|\mathcal{E}|$ is the number of entities in the KG \cite{dettmers2018}.


The \textit{fine-tuning optimization model} (Section~\ref{sec:fine-tuned-model}), in order to save on intensive computation, reuses the same hyper-parameters found for the KGE models. Depending on the optimizer choice, the choice of loss weights, $\alpha_C, \alpha_S,$ and $\alpha_{MLP}$, is important. However, our optimizer choice has dynamic learning rates per variable, and therefore, will adapt regardless of the loss weights and we can set $\alpha_C=\alpha_S=\alpha_{MLP}=~1$. 
Had we used, \eg stochastic gradient descent, these variables would needed to be tuned.

\subsubsection{Initialization of the fine-tuning optimization models}


As presented in Section~\ref{sec:fine-tuned-model}, we simultaneously train the KGE models and the MLP-based baseline model. This is done by initializing the model with \textit{(i)} the weights learned in the correspondent baseline model with pre-trained embeddings, and \textit{(ii)} the KG embeddings learned with the respective KGE models. For example, the \emph{Complex FT DistMult-HAKE} model is initialized with the learned weights with the \emph{Complex PT DistMult-HAKE} model and the pre-trained KG embeddings using DistMult and HAKE models.
Then the model is further trained with a small learning rate. We found that reducing the learning rate by a factor of $100$ worked well. Using this learning rate we optimize the model until convergence. 


\subsubsection{Simple and complex settings}
As presented in Section \ref{sec:baseline}, 
we use two settings in our classification models: simple and complex. This will help us isolate the effects of the KG embeddings versus the power of the MLP model.
The simple setting uses no branching layers, \ie $n_C=n_S=n_{\kappa}=0$ and $n=1$ as in Equations (\ref{eq:mlp2}), (\ref{eq:mlp3}), (\ref{eq:mlpkappa}) and \eqref{eq:mlp5} with $128$ units in the hidden dense layer. For the complex models we use random search (20 trials) to find the optimal number of layers and units out of the ranges shown in Table \ref{tab:hp}. The optimal choices for the top performing models (using one-hot and pre-trained embeddings) are shown in Table \ref{tab:pred_hp}. 

Looking at the increasing complexity of the layer configuration of the one-hot models in Table~\ref{tab:pred_hp} we can see a correlation from the simplest sampling strategy~(\ie~\textit{(i)}) through the most challenging one~(\ie~\textit{(iv)}). The same can be seen for 
PT HAKE-DisMult 
from strategy \textit{(iii)} to \textit{(iv)}, where the number of layers increase. 
Overall we can see that the layer configurations of the chemical branch is more complex than for the species branch. This indicates that the KGE models are better at representing $KG_S$ than~$KG_C$.

\subsection{Prediction results}
\label{sec:prediction_results}

In this section we present a summary of the conducted
chemical effect prediction evaluation. Complete results are available at the project repository.\footnote{\url{https://github.com/NIVA-Knowledge-Graph/KGs_and_Effect_Prediction_2020}}
The default decision threshold is set to $0.5$. That is, if a model predicts $\hat{y}>0.5$ for an input $x_{c,s,\kappa}$ then the chemical $c$ is considered lethal to $s$ at a concentration~$\kappa$.\footnote{\change{We set the decision threshold $\hat{y}>0.5$ since the model output bias (\cf Equation \eqref{eq:mlp6}) will be (close to) 0.5 after training. Recall that we have oversampled the classes to reach a $0.5/0.5$ prior probability during training (\cf Section \ref{sec:sampling-strategies}).}}

We use several metrics to compare the different prediction models. 
These are Sensitivity (\ie recall), Specificity, and Youden's index ($YI$) \cite{doi:10.1002/1097-0142(1950)3:1<32::AID-CNCR2820030106>3.0.CO;2-3}. \change{Precision and F-score were also considered as  metrics. However, they were not representative for the performance with respect to non-harmful chemicals. This is attributed to the larger number of positive samples (\ie harmful chemicals) than negative samples (\ie non-harmful chemicals) in the test data.
}

Sensitivity and Specificity are defined as 
\begin{align}
    \text{Sensitivity} =\frac{TP}{TP+FN},\\ \text{Specificity}= \frac{TN}{FP+TN},
\end{align}
where TP, FN, TN, and FP are true positives, false negatives, true negatives and false positives, respectively. 
YI is defined as 
\begin{align}
    YI = \text{Sensitivity} + \text{Specificity} - 1.
\end{align}
We also present the maximized Youden's index ($YI_{max}$), this is defined as
\begin{align}
    YI_{max} = \max_{\tau}\ \text{Sensitivity} + \text{Specificity} - 1,
\end{align}
\ie we maximize Youden's index based on the decision threshold ($\tau$), we call this optimal threshold $\tau_{max}$.
This metric is equivalent to the maximum of the Receiver operating characteristic (ROC) curve over a random model and can be used to select the optimal decision threshold in a production environment (based on validation data). We do not present ROC (or area under ROC, AUC) as a metric as it correlates ($>0.99$) with $YI_{max}$ in our case.


\change{In our setting, sensitivity is a measure on how well the models identify harmful chemicals while specificity measures models' ability to identify non-harmful chemicals.}
Youden's index is used to capture the usefulness of a diagnostic test (or in our case, a toxicity test). A useless test will have $YI=0$ while with $YI>0$ a test is useful. $YI$ is also thought of as how well informed a decision might be. 
Note that, $YI$ can be less than $0$, but this is solved by swapping labeled classes. Similarly to how negative correlation is still useful. 

\begin{table*}[]
    \centering
        \begin{tabular}{|l|c|c|c|c|c|}
        \hline 
\textbf{Model} & \textbf{Sensitivity} & \textbf{Specificity} & \textbf{YI} & \textbf{YI}$_{max}$ & $\mathbf{\tau}_{max}$\\ \hline \hline
Simple one-hot & $0.939 \pm 0.009$ & $0.657 \pm 0.018$ & $0.595 \pm 0.015$ & $0.666 \pm 0.011$ & $0.809 \pm 0.049$ \\ \hline\hline
Simple PT HAKE-HAKE & $0.912 \pm 0.006$ & $0.773 \pm 0.018$ & $0.685 \pm 0.016$ & $0.719 \pm 0.012$ & $0.707 \pm 0.044$ \\ \hline
Simple PT pRotatE-HAKE & $0.934 \pm 0.005$ & $0.749 \pm 0.044$ & $0.683 \pm 0.04$ & $0.718 \pm 0.02$ & $0.665 \pm 0.082$ \\ \hline
Simple PT ConvE-HAKE & $0.937 \pm 0.006$ & $0.738 \pm 0.006$ & $0.674 \pm 0.004$ & $0.724 \pm 0.007$ & $0.721 \pm 0.054$ \\ \hline
Simple PT pRotatE-ConvE & $0.924 \pm 0.029$ & $0.436 \pm 0.155$ & $0.36 \pm 0.182$ & $0.469 \pm 0.196$ & $0.784 \pm 0.052$ \\ \hline
Simple PT RotatE-ConvE & $\mathbf{0.997 \pm 0.003}$ & $0.024 \pm 0.035$ & $0.021 \pm 0.035$ & $0.195 \pm 0.111$ & $0.812 \pm 0.086$ \\ \hline\hline
Simple FT HAKE-HAKE & $0.921 \pm 0.005$ & ${0.814 \pm 0.009}$ & $\mathbf{0.734 \pm 0.006}$ & $0.743 \pm 0.007$ & $0.547 \pm 0.074$ \\ \hline
Simple FT pRotatE-HAKE & $0.92 \pm 0.005$ & $0.808 \pm 0.013$ & $\underline{0.728 \pm 0.011}$ & $0.738 \pm 0.007$ & $0.56 \pm 0.107$ \\ \hline
Simple FT ConvE-HAKE & $0.942 \pm 0.003$ & $0.733 \pm 0.019$ & $0.675 \pm 0.019$ & $0.729 \pm 0.007$ & $0.864 \pm 0.053$ \\ \hline
Simple FT pRotatE-ConvE & $0.949 \pm 0.003$ & $0.766 \pm 0.017$ & $0.715 \pm 0.016$ & $\mathbf{0.765 \pm 0.006}$ & $0.842 \pm 0.064$ \\ \hline
Simple FT RotatE-ConvE & $0.928 \pm 0.015$ & $0.797 \pm 0.036$ & $\underline{0.726 \pm 0.022}$ & $\underline{0.761 \pm 0.01}$ & $0.722 \pm 0.069$ \\ \hline\hline
Complex one-hot & $0.937 \pm 0.004$ & $0.748 \pm 0.016$ & $0.685 \pm 0.015$ & $0.728 \pm 0.009$ & $0.769 \pm 0.094$ \\ \hline\hline
Complex PT DistMult-HAKE & $0.895 \pm 0.008$ & ${0.817 \pm 0.008}$ & $0.713 \pm 0.007$ & $0.723 \pm 0.008$ & $0.456 \pm 0.088$ \\ \hline
Complex PT HAKE-ConvKB & $0.927 \pm 0.006$ & $0.784 \pm 0.017$ & $0.711 \pm 0.013$ & $0.739 \pm 0.009$ & $0.686 \pm 0.109$ \\ \hline
Complex PT HolE-ConvKB & $0.932 \pm 0.013$ & $0.779 \pm 0.024$ & $0.711 \pm 0.013$ & $0.729 \pm 0.009$ & $0.676 \pm 0.104$ \\ \hline
Complex PT ComplEx-DistMult & $0.96 \pm 0.006$ & $0.584 \pm 0.04$ & $0.543 \pm 0.039$ & $0.664 \pm 0.024$ & $0.838 \pm 0.048$ \\ \hline
Complex PT HolE-pRotatE & $\underline{0.996 \pm 0.006}$ & $0.011 \pm 0.02$ & $0.006 \pm 0.014$ & $0.182 \pm 0.041$ & $0.804 \pm 0.071$ \\ \hline\hline
Complex FT DistMult-HAKE & $0.903 \pm 0.009$ & ${0.816 \pm 0.015}$ & $0.719 \pm 0.008$ & $0.729 \pm 0.005$ & $0.597 \pm 0.098$ \\ \hline
Complex FT HAKE-ConvKB & $0.935 \pm 0.006$ & $0.791 \pm 0.021$ & $\underline{0.726 \pm 0.018}$ & ${0.754 \pm 0.008}$ & $0.776 \pm 0.109$ \\ \hline
Complex FT HolE-ConvKB & $0.895 \pm 0.01$ & $\mathbf{0.835 \pm 0.016}$ & $\underline{0.73 \pm 0.01}$ & $0.739 \pm 0.011$ & $0.61 \pm 0.123$ \\ \hline
Complex FT ComplEx-DistMult & $0.927 \pm 0.005$ & $0.78 \pm 0.018$ & $0.707 \pm 0.016$ & $0.742 \pm 0.011$ & $0.797 \pm 0.093$ \\ \hline
Complex FT HolE-pRotatE & $0.913 \pm 0.008$ & $0.795 \pm 0.017$ & $0.708 \pm 0.012$ & $0.734 \pm 0.008$ & $0.777 \pm 0.049$ \\ \hline
        \end{tabular} 
    \vspace{0.2cm}
    \caption{Prediction results (mean and standard deviation over 10 runs) for sampling strategy \textit{(i)}. 
    \textbf{Bold} denotes \textit{best mean result} and \underline{underline} denotes \textit{within one standard deviation of best result}.
    PT prefix denotes pre-trained and FT denotes fine-tuning. Simple denotes $n_C=n_S=n_{\kappa}=0$ and $n=1$ while in complex, $n_C,n_S,n_{\kappa}$ and $n$ are hyper-parameters in Equations (\ref{eq:mlp2}), (\ref{eq:mlp3}), (\ref{eq:mlpkappa}) and \eqref{eq:mlp5}.
    }
    \label{tab:results_1}
\end{table*}

\begin{table*}[]
    \centering
        \begin{tabular}{|l|c|c|c|c|c|}
        \hline 
\textbf{Model} & \textbf{Sensitivity} & \textbf{Specificity} & \textbf{YI} & \textbf{YI}$_{max}$ & $\mathbf{\tau}_{max}$\\ \hline\hline 
Simple one-hot & $0.88 \pm 0.022$ & $0.628 \pm 0.048$ & $0.508 \pm 0.057$ & $0.556 \pm 0.051$ & $0.713 \pm 0.13$ \\ \hline\hline 
Simple PT HAKE-ConvKB & $0.926 \pm 0.007$ & $0.823 \pm 0.016$ & $0.748 \pm 0.017$ & $0.775 \pm 0.013$ & $0.623 \pm 0.064$ \\ \hline
Simple PT HAKE-HAKE & $0.908 \pm 0.007$ & $0.829 \pm 0.014$ & $0.738 \pm 0.012$ & $0.759 \pm 0.01$ & $0.613 \pm 0.132$ \\ \hline
Simple PT pRotatE-HAKE & $0.924 \pm 0.003$ & $0.802 \pm 0.009$ & $0.726 \pm 0.008$ & $0.76 \pm 0.006$ & $0.79 \pm 0.084$ \\ \hline
Simple PT RotatE-ConvKB & $0.972 \pm 0.021$ & $0.42 \pm 0.255$ & $0.392 \pm 0.236$ & $0.62 \pm 0.111$ & $0.814 \pm 0.06$ \\ \hline
Simple PT RotatE-ConvE & $\mathbf{0.997 \pm 0.004}$ & $0.021 \pm 0.057$ & $0.018 \pm 0.054$ & $0.22 \pm 0.088$ & $0.824 \pm 0.095$ \\ \hline\hline 
Simple FT HAKE-ConvKB & $0.909 \pm 0.003$ & $0.883 \pm 0.006$ & $\mathbf{0.792 \pm 0.006}$ & ${0.803 \pm 0.004}$ & $0.556 \pm 0.138$ \\ \hline
Simple FT HAKE-HAKE & $0.897 \pm 0.007$ & $0.86 \pm 0.01$ & $0.757 \pm 0.012$ & $0.769 \pm 0.006$ & $0.61 \pm 0.134$ \\ \hline
Simple FT pRotatE-HAKE & $0.905 \pm 0.004$ & $0.859 \pm 0.012$ & $0.764 \pm 0.012$ & $0.775 \pm 0.011$ & $0.544 \pm 0.099$ \\ \hline
Simple FT RotatE-ConvKB & $0.93 \pm 0.007$ & $0.853 \pm 0.013$ & $\underline{0.784 \pm 0.008}$ & $\mathbf{0.81 \pm 0.008}$ & $0.732 \pm 0.119$ \\ \hline
Simple FT RotatE-ConvE & $0.912 \pm 0.02$ & $0.821 \pm 0.028$ & $0.733 \pm 0.01$ & $0.753 \pm 0.005$ & $0.735 \pm 0.17$ \\ \hline\hline 
Complex one-hot & $0.875 \pm 0.014$ & $0.859 \pm 0.015$ & $0.734 \pm 0.012$ & $0.749 \pm 0.009$ & $0.448 \pm 0.2$ \\ \hline\hline 
Complex PT HolE-ConvKB & $0.894 \pm 0.006$ & ${0.889 \pm 0.014}$ & $\underline{0.783 \pm 0.014}$ & ${0.793 \pm 0.01}$ & $0.489 \pm 0.035$ \\ \hline
Complex PT pRotatE-ConvKB & $0.901 \pm 0.012$ & $0.875 \pm 0.027$ & $\underline{0.776 \pm 0.024}$ & $0.79 \pm 0.018$ & $0.592 \pm 0.081$ \\ \hline
Complex PT TransE-ConvKB & $0.906 \pm 0.008$ & $0.868 \pm 0.021$ & $\underline{0.774 \pm 0.019}$ & $0.787 \pm 0.012$ & $0.588 \pm 0.112$ \\ \hline
Complex PT ComplEx-ConvE & $0.928 \pm 0.006$ & $0.768 \pm 0.015$ & $0.696 \pm 0.015$ & $0.731 \pm 0.008$ & $0.689 \pm 0.095$ \\ \hline
Complex PT ConvKB-pRotatE & $\underline{0.995 \pm 0.005}$ & $0.011 \pm 0.012$ & $0.007 \pm 0.008$ & $0.265 \pm 0.054$ & $0.77 \pm 0.089$ \\ \hline\hline 
Complex FT HolE-ConvKB & $0.871 \pm 0.007$ & ${0.906 \pm 0.007}$ & ${0.778 \pm 0.007}$ & $0.791 \pm 0.005$ & $0.441 \pm 0.07$ \\ \hline
Complex FT pRotatE-ConvKB & $0.869 \pm 0.008$ & $\mathbf{0.914 \pm 0.011}$ & ${0.783 \pm 0.007}$ & $0.794 \pm 0.006$ & $0.483 \pm 0.083$ \\ \hline
Complex FT TransE-ConvKB & $0.878 \pm 0.008$ & ${0.895 \pm 0.011}$ & $0.772 \pm 0.008$ & $0.792 \pm 0.006$ & $0.511 \pm 0.133$ \\ \hline
Complex FT ComplEx-ConvE & $0.916 \pm 0.009$ & $0.83 \pm 0.021$ & $0.746 \pm 0.016$ & $0.76 \pm 0.011$ & $0.596 \pm 0.151$ \\ \hline
Complex FT ConvKB-pRotatE & $0.9 \pm 0.013$ & $0.794 \pm 0.026$ & $0.694 \pm 0.018$ & $0.723 \pm 0.014$ & $0.785 \pm 0.111$ \\ \hline
        \end{tabular} 
    \vspace{0.2cm}
    \caption{Prediction results for sampling strategy \textit{(ii)}. Same notation as Table \ref{tab:results_1}.
    }
    \label{tab:results_2}
\end{table*}
\begin{table*}[]
    \centering
        \begin{tabular}{|l|c|c|c|c|c|}
        \hline 
\textbf{Model} & \textbf{Sensitivity} & \textbf{Specificity} & \textbf{YI} & \textbf{YI}$_{max}$ & $\mathbf{\tau}_{max}$\\\hline\hline
Simple one-hot & $0.822 \pm 0.058$ & $0.439 \pm 0.054$ & $0.261 \pm 0.058$ & $0.31 \pm 0.047$ & $0.597 \pm 0.182$ \\ \hline\hline
Simple PT ConvKB-DistMult & $0.966 \pm 0.007$ & ${0.626 \pm 0.047}$ & $\underline{0.591 \pm 0.045}$ & $\underline{0.623 \pm 0.049}$ & $0.67 \pm 0.058$ \\ \hline
Simple PT HAKE-DistMult & $0.958 \pm 0.023$ & $0.628 \pm 0.026$ & ${0.586 \pm 0.033}$ & $\underline{0.626 \pm 0.045}$ & $0.613 \pm 0.092$ \\ \hline
Simple PT ConvKB-TransE & $0.969 \pm 0.009$ & $0.614 \pm 0.048$ & $\underline{0.583 \pm 0.04}$ & $\underline{0.642 \pm 0.01}$ & $0.643 \pm 0.059$ \\ \hline
Simple PT ConvE-RotatE & $0.934 \pm 0.055$ & $0.276 \pm 0.026$ & $0.209 \pm 0.043$ & $0.273 \pm 0.071$ & $0.596 \pm 0.13$ \\ \hline
Simple PT HolE-HAKE & $0.88 \pm 0.089$ & $0.115 \pm 0.083$ & $-0.005 \pm 0.075$ & $0.077 \pm 0.057$ & $0.783 \pm 0.18$ \\ \hline\hline
Simple FT ConvKB-DistMult & $0.947 \pm 0.014$ & ${0.667 \pm 0.02}$ & $\underline{0.614 \pm 0.013}$ & $\underline{0.645 \pm 0.011}$ & $0.736 \pm 0.087$ \\ \hline
Simple FT HAKE-DistMult & $0.947 \pm 0.012$ & $\underline{0.662 \pm 0.035}$ & $\underline{0.609 \pm 0.031}$ & $\underline{0.634 \pm 0.026}$ & $0.701 \pm 0.132$ \\ \hline
Simple FT ConvKB-TransE & $0.934 \pm 0.009$ & $\underline{0.68 \pm 0.018}$ & $\underline{0.615 \pm 0.014}$ & $\underline{0.642 \pm 0.015}$ & $0.687 \pm 0.065$ \\ \hline
Simple FT ConvE-RotatE & $0.915 \pm 0.013$ & $0.454 \pm 0.028$ & $0.369 \pm 0.027$ & $0.402 \pm 0.028$ & $0.658 \pm 0.083$ \\ \hline
Simple FT HolE-HAKE & $0.931 \pm 0.009$ & $0.118 \pm 0.036$ & $0.049 \pm 0.038$ & $0.171 \pm 0.038$ & $0.882 \pm 0.127$ \\ \hline\hline
Complex one-hot & $0.796 \pm 0.028$ & $0.571 \pm 0.041$ & $0.367 \pm 0.054$ & $0.398 \pm 0.043$ & $0.526 \pm 0.076$ \\ \hline\hline
Complex PT HAKE-DistMult & ${0.969 \pm 0.016}$ & $\underline{0.642 \pm 0.044}$ & $\underline{0.61 \pm 0.034}$ & $\underline{0.643 \pm 0.026}$ & $0.675 \pm 0.105$ \\ \hline
Complex PT pRotatE-ComplEx & $0.929 \pm 0.024$ & $\underline{0.668 \pm 0.048}$ & $\underline{0.597 \pm 0.048}$ & $\underline{0.62 \pm 0.046}$ & $0.526 \pm 0.145$ \\ \hline
Complex PT ConvKB-DistMult & $0.965 \pm 0.013$ & $\underline{0.631 \pm 0.078}$ & $\underline{0.597 \pm 0.07}$ & $\underline{0.627 \pm 0.039}$ & $0.597 \pm 0.149$ \\ \hline
Complex PT ComplEx-HolE & $\mathbf{0.991 \pm 0.01}$ & $0.237 \pm 0.106$ & $0.228 \pm 0.098$ & $0.45 \pm 0.028$ & $0.721 \pm 0.047$ \\ \hline
Complex PT ComplEx-HAKE & $0.9 \pm 0.055$ & $0.097 \pm 0.047$ & $-0.003 \pm 0.064$ & $0.133 \pm 0.081$ & $0.696 \pm 0.22$ \\ \hline\hline
Complex FT HAKE-DistMult & $0.932 \pm 0.011$ & $\mathbf{0.69 \pm 0.024}$ & $\mathbf{0.622 \pm 0.023}$ & $\mathbf{0.652 \pm 0.022}$ & $0.706 \pm 0.134$ \\ \hline
Complex FT pRotatE-ComplEx & $0.931 \pm 0.025$ & $\underline{0.672 \pm 0.042}$ & $\underline{0.602 \pm 0.045}$ & $\underline{0.631 \pm 0.037}$ & $0.627 \pm 0.157$ \\ \hline
Complex FT ConvKB-DistMult & $0.953 \pm 0.008$ & ${0.642 \pm 0.027}$ & $\underline{0.596 \pm 0.027}$ & $\underline{0.625 \pm 0.028}$ & $0.753 \pm 0.138$ \\ \hline
Complex FT ComplEx-HolE & $0.898 \pm 0.035$ & $0.591 \pm 0.064$ & $0.489 \pm 0.042$ & $0.521 \pm 0.027$ & $0.612 \pm 0.156$ \\ \hline
Complex FT ComplEx-HAKE & $0.88 \pm 0.032$ & $0.255 \pm 0.026$ & $0.135 \pm 0.034$ & $0.204 \pm 0.06$ & $0.775 \pm 0.268$ \\ \hline
        \end{tabular} 
    \vspace{0.2cm}
    \caption{Prediction results for sampling strategy \textit{(iii)}. Same notation as Table \ref{tab:results_1}.
    }
    \label{tab:results_3}
\end{table*}
\begin{table*}[]
    \centering
        \begin{tabular}{|l|c|c|c|c|c|}
        \hline 
\textbf{Model} & \textbf{Sensitivity} & \textbf{Specificity} & \textbf{YI} & \textbf{YI}$_{max}$ & $\mathbf{\tau}_{max}$\\ \hline\hline
Simple one-hot & $0.612 \pm 0.096$ & $0.421 \pm 0.107$ & $0.033 \pm 0.14$ & $0.113 \pm 0.076$ & $0.555 \pm 0.306$ \\ \hline\hline
Simple PT HAKE-ComplEx & $\underline{0.971 \pm 0.011}$ & $0.361 \pm 0.065$ & $0.332 \pm 0.056$ & $\underline{0.546 \pm 0.031}$ & $0.89 \pm 0.042$ \\ \hline
Simple PT pRotatE-ComplEx & $\mathbf{0.972 \pm 0.008}$ & $0.36 \pm 0.079$ & $0.332 \pm 0.074$ & $\underline{0.527 \pm 0.045}$ & $0.852 \pm 0.04$ \\ \hline
Simple PT HolE-ComplEx & $\underline{0.965 \pm 0.032}$ & $0.363 \pm 0.068$ & $0.328 \pm 0.063$ & $\underline{0.549 \pm 0.075}$ & $0.856 \pm 0.077$ \\ \hline
Simple PT pRotatE-RotatE & $0.917 \pm 0.01$ & $0.168 \pm 0.016$ & $0.084 \pm 0.013$ & $0.151 \pm 0.021$ & $0.779 \pm 0.182$ \\ \hline
Simple PT HAKE-HAKE & $0.8 \pm 0.095$ & $0.128 \pm 0.066$ & $-0.072 \pm 0.07$ & $0.033 \pm 0.027$ & $0.736 \pm 0.321$ \\ \hline\hline
Simple FT HAKE-ComplEx & $\underline{0.963 \pm 0.01}$ & $0.423 \pm 0.102$ & $\underline{0.386 \pm 0.096}$ & $\underline{0.57 \pm 0.03}$ & $0.875 \pm 0.079$ \\ \hline
Simple FT pRotatE-ComplEx & $0.954 \pm 0.009$ & $\underline{0.5 \pm 0.058}$ & $\underline{0.454 \pm 0.052}$ & $\underline{0.569 \pm 0.024}$ & $0.854 \pm 0.073$ \\ \hline
Simple FT HolE-ComplEx & $\underline{0.965 \pm 0.007}$ & $0.418 \pm 0.058$ & ${0.383 \pm 0.053}$ & $\mathbf{0.571 \pm 0.042}$ & $0.9 \pm 0.046$ \\ \hline
Simple FT pRotatE-RotatE & $0.806 \pm 0.039$ & $0.229 \pm 0.027$ & $0.035 \pm 0.016$ & $0.131 \pm 0.032$ & $0.782 \pm 0.157$ \\ \hline
Simple FT HAKE-HAKE & $0.893 \pm 0.046$ & $0.104 \pm 0.051$ & $-0.003 \pm 0.031$ & $0.037 \pm 0.033$ & $0.588 \pm 0.332$ \\ \hline\hline
Complex one-hot & $0.656 \pm 0.069$ & $0.422 \pm 0.075$ & $0.078 \pm 0.053$ & $0.124 \pm 0.036$ & $0.645 \pm 0.178$ \\ \hline\hline
Complex PT HAKE-DistMult & $0.923 \pm 0.013$ & $0.434 \pm 0.059$ & $0.357 \pm 0.052$ & ${0.488 \pm 0.074}$ & $0.808 \pm 0.07$ \\ \hline
Complex PT HolE-DistMult & ${0.949 \pm 0.016}$ & $0.38 \pm 0.084$ & $0.33 \pm 0.076$ & $0.443 \pm 0.089$ & $0.805 \pm 0.07$ \\ \hline
Complex PT ConvKB-DistMult & $0.942 \pm 0.01$ & $0.387 \pm 0.038$ & $0.329 \pm 0.039$ & ${0.484 \pm 0.066}$ & $0.817 \pm 0.052$ \\ \hline
Complex PT HolE-RotatE & $0.932 \pm 0.014$ & $0.15 \pm 0.018$ & $0.082 \pm 0.023$ & $0.168 \pm 0.015$ & $0.861 \pm 0.064$ \\ \hline
Complex PT TransE-HAKE & $0.756 \pm 0.047$ & $0.19 \pm 0.077$ & $-0.054 \pm 0.089$ & $0.057 \pm 0.046$ & $0.742 \pm 0.253$ \\ \hline\hline
Complex FT HAKE-DistMult & $0.925 \pm 0.021$ & $\underline{0.513 \pm 0.064}$ & $\underline{0.437 \pm 0.058}$ & ${0.522 \pm 0.034}$ & $0.83 \pm 0.09$ \\ \hline
Complex FT HolE-DistMult & $0.926 \pm 0.015$ & $\mathbf{0.536 \pm 0.03}$ & $\mathbf{0.462 \pm 0.03}$ & $\underline{0.543 \pm 0.039}$ & $0.81 \pm 0.084$ \\ \hline
Complex FT ConvKB-DistMult & $0.933 \pm 0.01$ & $\underline{0.525 \pm 0.065}$ & $\underline{0.459 \pm 0.063}$ & $\underline{0.55 \pm 0.04}$ & $0.746 \pm 0.122$ \\ \hline
Complex FT HolE-RotatE & $0.863 \pm 0.057$ & $0.194 \pm 0.053$ & $0.057 \pm 0.015$ & $0.11 \pm 0.021$ & $0.81 \pm 0.278$ \\ \hline
Complex FT TransE-HAKE & $0.892 \pm 0.027$ & $0.075 \pm 0.043$ & $-0.033 \pm 0.049$ & $0.072 \pm 0.048$ & $0.958 \pm 0.077$ \\ \hline
        \end{tabular} 
    \vspace{0.2cm}
    \caption{Prediction results sampling strategy \textit{(iv)}. Same notation as Table \ref{tab:results_1}.
    }
    \label{tab:results_4}
\end{table*}

Tables \ref{tab:results_1}-\ref{tab:results_4} 
show the results for each of the data sampling strategies \textit{(i)-(iv)}, respectively.
The tables include the three best models (based on $YI$) for the baseline model using one-hot and pre-trained (PT) KG embeddings, and the fine-tuning (FT)  models using the same combination of KGE models as the selected PT-based models.
We have also included a model with middling performance 
(\ie 40 out of 81 models) and the worst performing model. Note that for the PT- and FT-based models we have evaluated 81 combinations KGE$_C$-KGE$_S$ of KGE models. 
All models were evaluated using the simple and complex MLP settings. For example, the model \emph{Complex FT DistMult-HolE} denotes that fine-tuning was used together with the complex MLP setting, and DistMult was selected to embed the chemicals $KG_C$ while HolE was used to embed the species $KG_S$.
We present the mean and standard deviation over $10$ evaluation runs, \ie we re-initialize and re-train the models $10$ times. Results highlighted in \textbf{bold} 
are the best mean
results of the corresponding metrics.
\underline{Underlined} results are where there is a $\geq 32\%$ chance that a single run outperforms the best mean (\ie one standard deviation contains $68\%$ of results, assuming normally distribute results).\footnote{Note that we only consider the best mean result and not the standard deviation in both directions.}

Overall, 
models with the complex setting
and fine-tuning 
are needed as the data sampling strategies become more challenging. Moreover, all models favour sensitivity over specificity at default decision threshold ($0.5$). This is down to the imbalance in the data. We can see the imbalance by $\tau_{max}$, it is $>0.5$ for most models. 
As we use a log-loss instead of a discrete loss, this is to be expected for imbalanced data. 

For settings \textit{(iii)} and \textit{(iv)} the performance drops and the standard deviation increases compared to the other strategies. This large standard deviation leads to large overlaps in quantiles among top-3 models in all categories, such that, by chance, one of these models could perform best in one individual evaluation.

\subsubsection{One-hot baseline models}
For the sampling strategy \textit{(i)} the one-hot baseline models perform well, especially, with the complex one-hot model. This complex model is equivalent in terms of $YI$ as the best simple pre-trained model. The story is largely the same in setting \textit{(ii)}, where the complex one-hot model performs within $1.5\%$ of the best simple pre-trained models. With strategies \textit{(iii)} and \textit{(iv)} the one-hot models degrade, especially in strategy \textit{(iv)} where the Youden's index is near zero \change{($<0.1$)}. This is expected as the one-hot baseline models lack important background information about the entities, specially for unseen chemicals and species, that the KG embedding models aim at capturing.

\subsubsection{Baseline with pre-trained KG embeddings}
We can see that the PT-based models 
do not lead to an important improvement with respect to
$YI_{max}$ in sampling strategy \textit{(i)}. 
The top-1 complex PT model, however, yields a better balance between sensitivity and specificity leading to an improved $YI$ over the complex one-hot models. The two middling performing models, \textit{Simple PT pRotatE-ConvE} and \textit{Complex PT ComplEx-DistMult}, still retain a decent level of performance.

The results with the strategy  \textit{(ii)} are similar to strategy \textit{(i)}, the delta in $YI$ between the simple and the complex PT-based models are about $5\%$. This slight improvement is due to the increased balance between sensitivity and specificity which in turn leads to a higher $YI$. 

In the sampling strategy \textit{(iii)} we can observe that the improvement of the PT-based models over the one-hot models increases. The increase is up to $25\%$ in $YI$ of the the best PT-based model over the best one-hot model. In addition, we observe in this strategy that the standard deviation increases, especially in specificity, leading to a large portion of the models that are within one standard deviation of the best model in terms of~$YI$.

Finally, the impact of using a PT-based models is strengthen in strategy \textit{(iv)}. The delta between the one-hot and PT-based models is up to $40\%$ in $YI$, and larger for $YI_{max}$. We see that all models struggle with specificity in this setting, this is down to the difficulty of predicting true negatives. This also leads to a larger variation, with certain models yielding standard deviation in the same order of magnitude as the metric (\eg \textit{Simple FT HAKE-ComplEx}).

\subsubsection{Fine-tuning optimization model}
The FT-based models, with some exceptions, improve the results over the PT-based models, most notably in sampling strategies \textit{(iii)} and \textit{(iv)}.
For example, the FT-based models \textit{Complex FT HolE-DistMult} and \textit{Simple FT HolE-ComplEx} are the best models in terms of $YI$ and $YI_{max}$ in strategy \textit{(iv)}, respectively.
We can also see in strategies \textit{(i)} and \textit{(ii)} that the FT-based models 
improve middling and worst performing PT-based models, \eg \textit{Simple FT RotatE-ConvE} in strategy \textit{(i)} improves from $YI=0.021$ to $YI=0.726$ using fine-tuning of the KG embeddings. 
The results are expected as the fine-tuned KG embeddings are tailored to the effect prediction~task.

\subsection{KG embedding analysis}
In this section we look at correlations between KGE model choices and prediction performance. KGE models are designed to capture certain structures in the data, and this can give some explanation of which parts of the KGs are important for prediction.

First, in Table \ref{tab:best_models} we show how many times a KGE model is used when regarding the top 10 performing combinations (out of the total 81 possible combinations). We focus on the choices when using the simple MLP setting 
to reduce the influence of the non-linear transforms on the embeddings. 

\begin{table*}[]
    \centering
    \begin{tabular}{|l|c|c|c|c|}\hline 
         \textbf{KGE model} & \textbf{\# uses \textit{(i)}}& \textbf{\# uses \textit{(ii)}}& \textbf{\# uses \textit{(iii)}} & \textbf{\# uses \textit{(iv)}}\\\hline\hline
        DistMult & $1/0$ & $0/1$ & $1/7$ & $0/4$ \\ \hline
        ComplEx & $1/1$ & $1/3$ & $2/1$ & $1/5$ \\ \hline
        HolE & $2/0$ & $1/0$ & $1/0$ & $1/0$ \\ \hline
        \textbf{Total decomposition} & $4/1$ & $2/4$ & $4/8$ & $2/9$ \\ \hline\hline 
        TransE & $1/0$ & $2/0$ & $1/2$ & $0/0$ \\ \hline
        RotatE & $0/0$ & $0/0$ & $0/0$ & $1/0$ \\ \hline
        pRotatE & $1/0$ & $1/0$ & $1/0$ & $3/0$ \\ \hline
        HAKE & $2/8$ & $3/5$ & $1/0$ & $2/0$ \\ \hline
        \textbf{Total geometric} & $4/8$ & $6/5$ & $3/2$ & $5/0$ \\ \hline\hline 
        ConvKB & $1/1$ & $0/1$ & $2/0$ & $0/1$ \\ \hline
        ConvE & $1/0$ & $2/0$ & $1/0$ & $2/0$ \\ \hline
        \textbf{Total convolutional} & $2/1$ & $2/1$ & $3/0$ & $2/1$ \\ \hline
    \end{tabular}
    \vspace{0.2cm}
    \caption{Usage of KGE models for each sampling strategy in simple MLP setting in top-10 performing combinations. Note that, there is one model for the $KG_C$ and one for $KG_S$, such that there is a total of 20 models per sampling strategy. 
    Notation: `\text{used} in $KG_C$ /  \text{used} in $KG_S$', \eg HAKE, $2/8$ in sampling strategy \textit{(i)}, indicates that HAKE is used to embed $KG_C$ 2 out of top-10 combinations and it is used to embed $KG_S$ 8 out of top-10 combinations. 
    }
    \label{tab:best_models}
\end{table*}

Looking at Table \ref{tab:best_models} we can see that the KGE models used to embed the chemicals $KG_C$ in the best performing models is distributed evenly across most models and settings. This indicates that the performance of the prediction models is not highly correlated with the use of a KGE model on $KG_C$. \change{Referencing Table \ref{tab:stats}, the high relational density in $KG_C$ can contribute to worse performance \cite{pujara-etal-2017-sparsity} and therefore equal distribution of models in Table \ref{tab:best_models}.}
This is different for $KG_S$. For sampling strategies \textit{(i)} and \textit{(ii)}, HAKE is extensively used in the top models to embed $KG_S$. HAKE is designed to embed hierarchies. Therefore, this indicates that in strategies \textit{(i)} and \textit{(ii)} the hierarchical structure of $KG_S$ dwarfs the rest of the KG. \change{$KG_S$ has a higher entity density and lower entity entropy (Table \ref{tab:stats}) than $KG_C$. This should lead to higher performance generally, but might also lead to larger discrepancies between models as seen in Table \ref{tab:best_models}.}

\begin{figure*}[t]
     \centering 
     \begin{subfigure}[t]{0.47\textwidth}
         \centering
         \includegraphics[width=\textwidth]{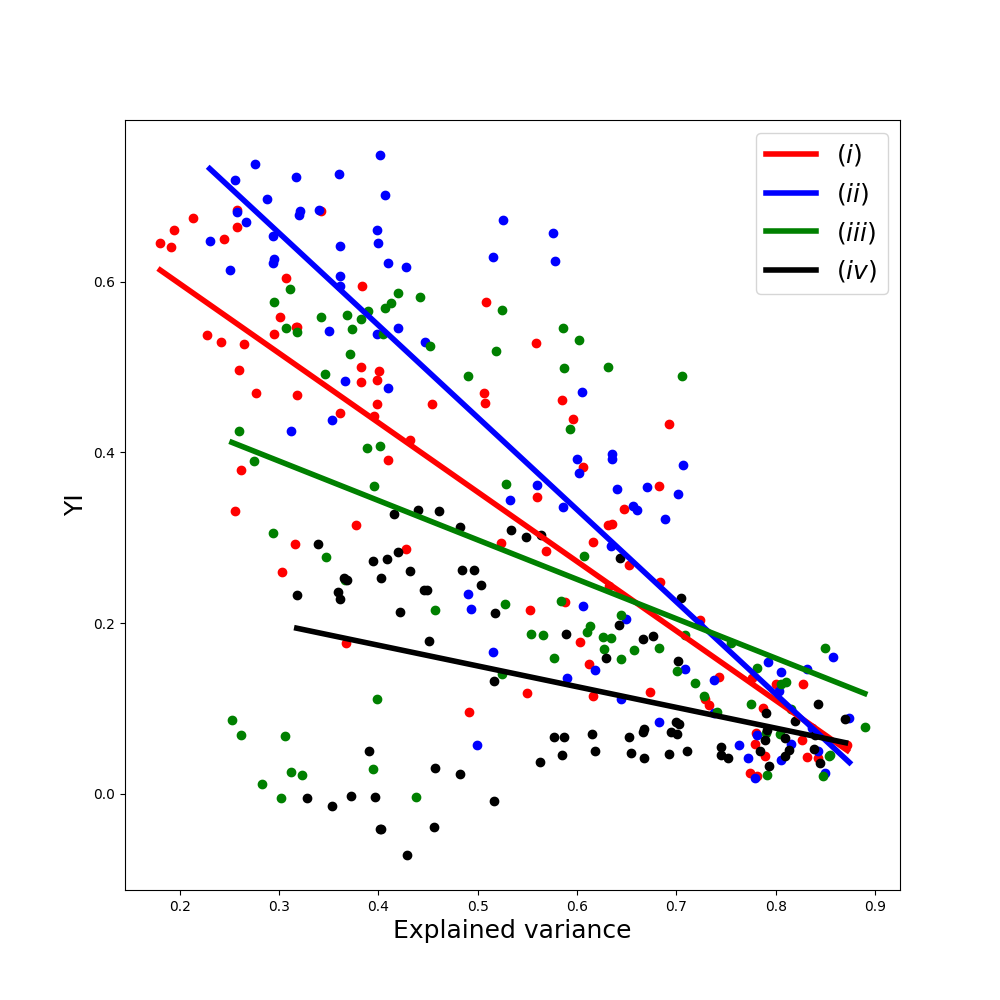}
         \caption{Simple PT models.}
         \label{fig:EVvsYI_simple}
     \end{subfigure}
     \hfill
     \begin{subfigure}[t]{0.47\textwidth}
         \centering
         \includegraphics[width=\textwidth]{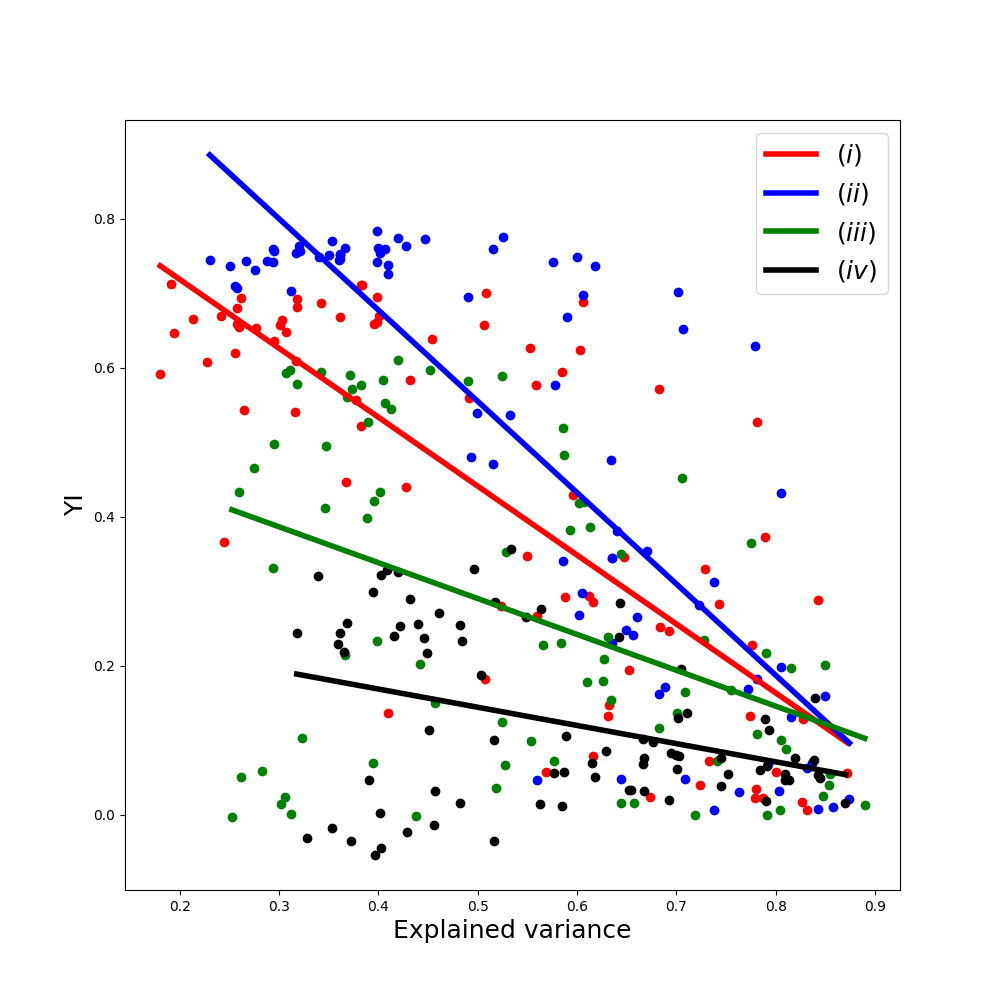}
         \caption{Complex PT models.}
         \label{fig:EVvsYI_complex}
     \end{subfigure}
        \caption{Relation between explained variance using 10 principal components and model performance represented as $YI$.}
        \label{fig:EVvsYI}
\end{figure*}

The use of the decomposition models increase in strategies \textit{(iii)} and \textit{(iv)} for the embedding of $KG_S$, which indicates that KG structures, other
than the hierarchy, are important. 
Overall, DistMult and ComplEx can be used to great effect in strategies \textit{(iii)} and \textit{(iv)} while the geometric model, HAKE, is more successful in the less challenging strategies \textit{(i)} and \textit{(ii)}.

\begin{figure*}[t]
     \centering 
     \begin{subfigure}[t]{0.47\textwidth}
         \centering
         \includegraphics[width=\textwidth]{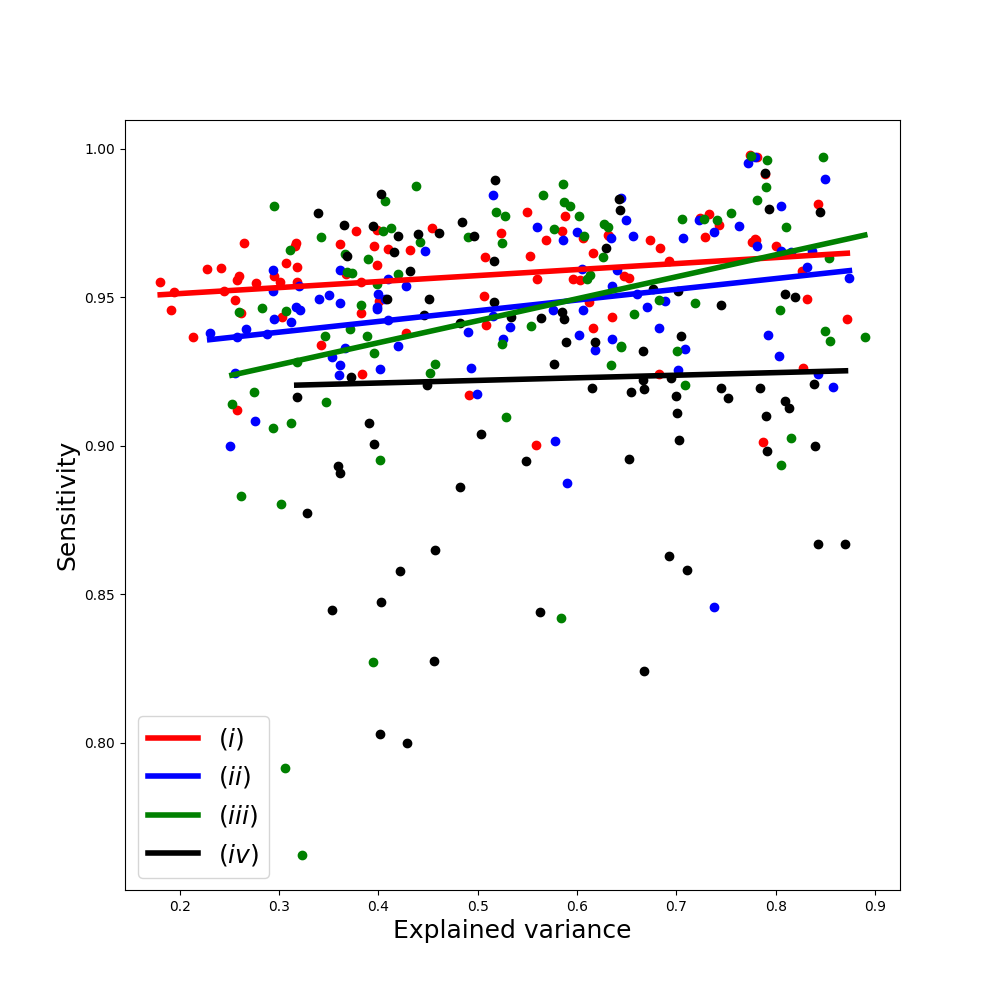}
          \caption{Simple PT models.}
         \label{fig:EVvsSensitivity_simple}
     \end{subfigure}
     \hfill
     \begin{subfigure}[t]{0.47\textwidth}
         \centering
         \includegraphics[width=\textwidth]{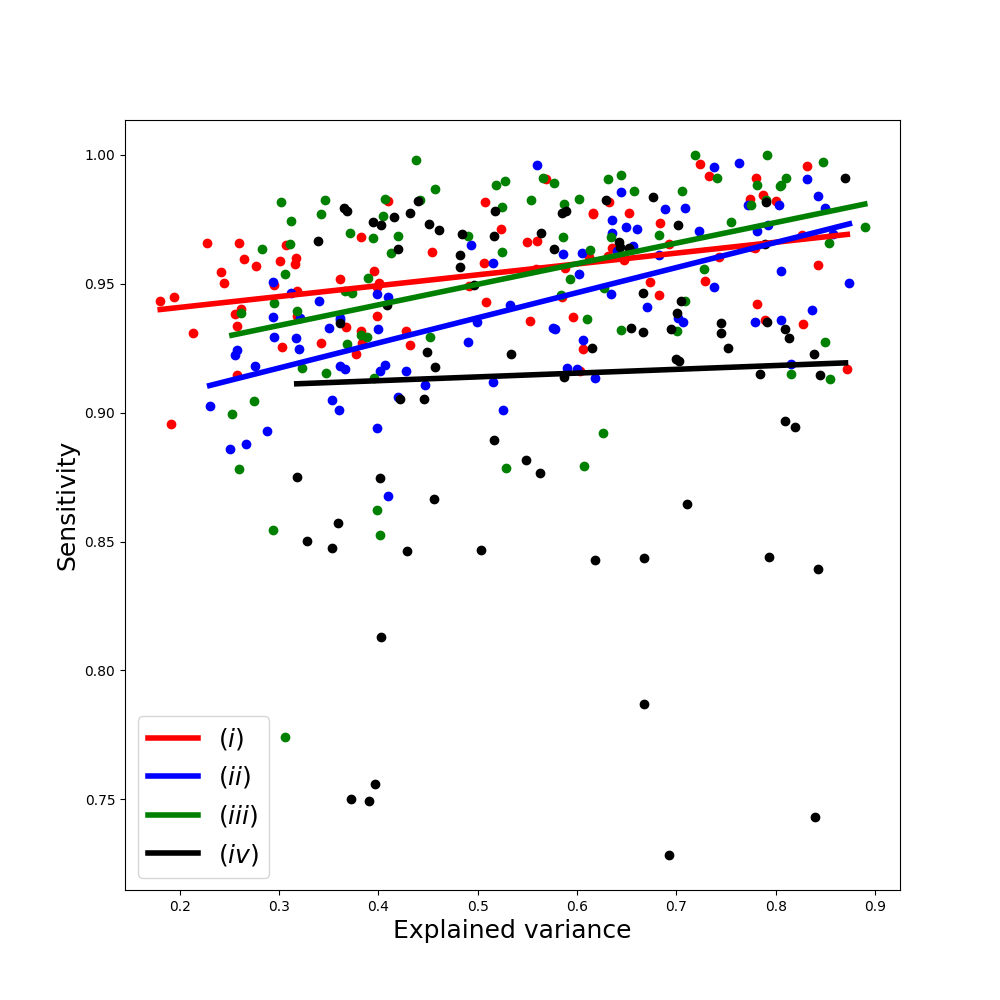}
         \caption{Complex PT models.}
         \label{fig:EVvsSensitivity_complex}
     \end{subfigure}
        \caption{Relation between explained variance using 10 principal components and model performance represented as sensitivity.}
        \label{fig:EVvsSensitivity}
\end{figure*}

\subsubsection{Explained variance}


Explained variance is a measure of how many principal components 
are required to describe all components.\footnote{We use the scikit-learn implementation \cite{scikit-learn} based on \cite{10.2307/2680726}.} In Figure \ref{fig:EVvsYI}, we present how the $YI$ metric depends on the explained variance of the top-10 principal components (\ie $\sum_{i=1}^{10} pca_i$). We show all (81 per sampling strategy) PT-based prediction model results, simple MLP setting in Figure \ref{fig:EVvsYI_simple} and complex setting in Figure \ref{fig:EVvsYI_complex}. For example, in Figure \ref{fig:EVvsYI_simple}, the best model in the strategy \textit{(iv)}, \textit{Simple PT pRotatE-ComplEx} have a explained variance of $0.49$ compared to the worst model, \textit{Simple PT HAKE-HAKE}, with explained variance of $0.34$. Coincidentally, these two points does not follow the trend lines in these figures which indicate negative correlation between $YI$ and explained variance. The trend lines can be interpreted in two ways. First, it is counter-intuitive as we would expect more descriptive embeddings, \ie larger explained variance, to perform better. On the other hand, the top-10 principal components may not be representative enough to capture the semantics of the KG embeddings, and thus, a large 
explained variance does not necessarily correlate with a high performance.

\begin{figure*}[t]
     \centering 
     \begin{subfigure}[t]{0.47\textwidth}
         \centering
         \includegraphics[width=\textwidth]{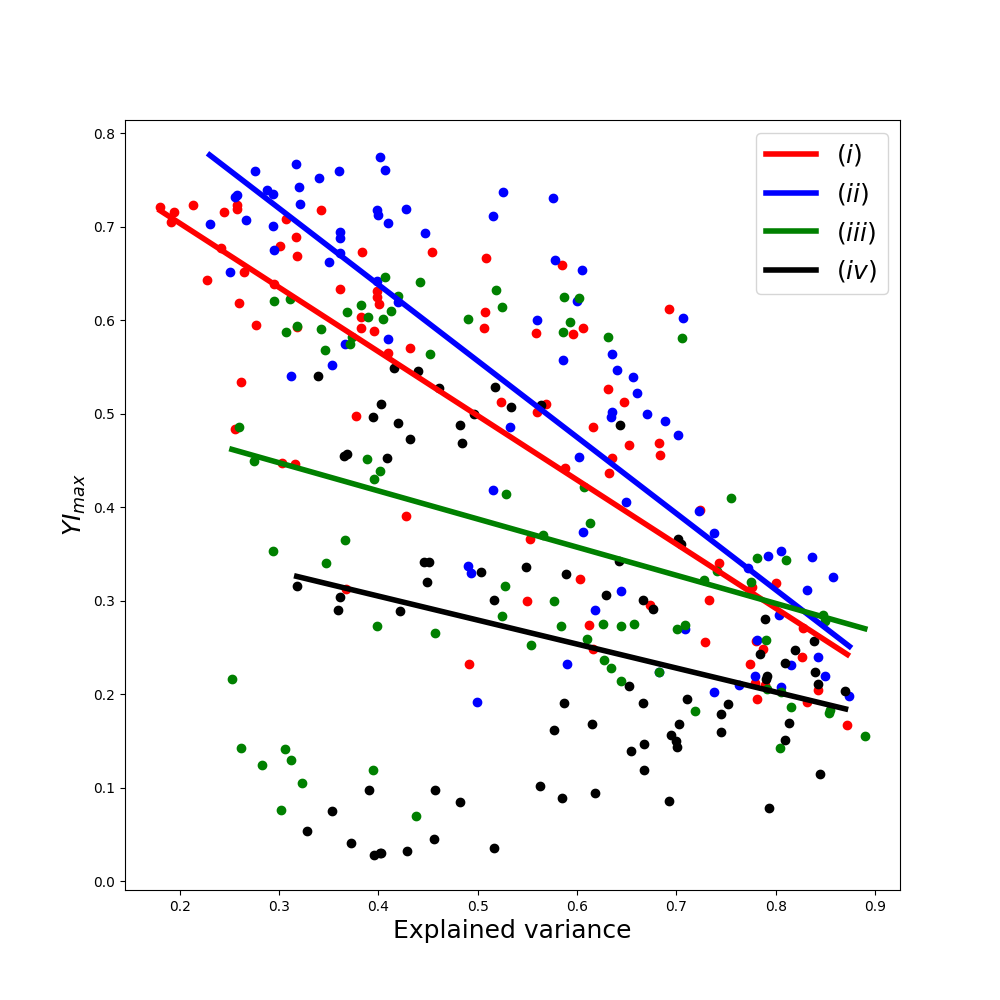}
          \caption{Simple PT models.}
         \label{fig:EVvsYImax_simple}
     \end{subfigure}
     \hfill
     \begin{subfigure}[t]{0.47\textwidth}
         \centering
         \includegraphics[width=\textwidth]{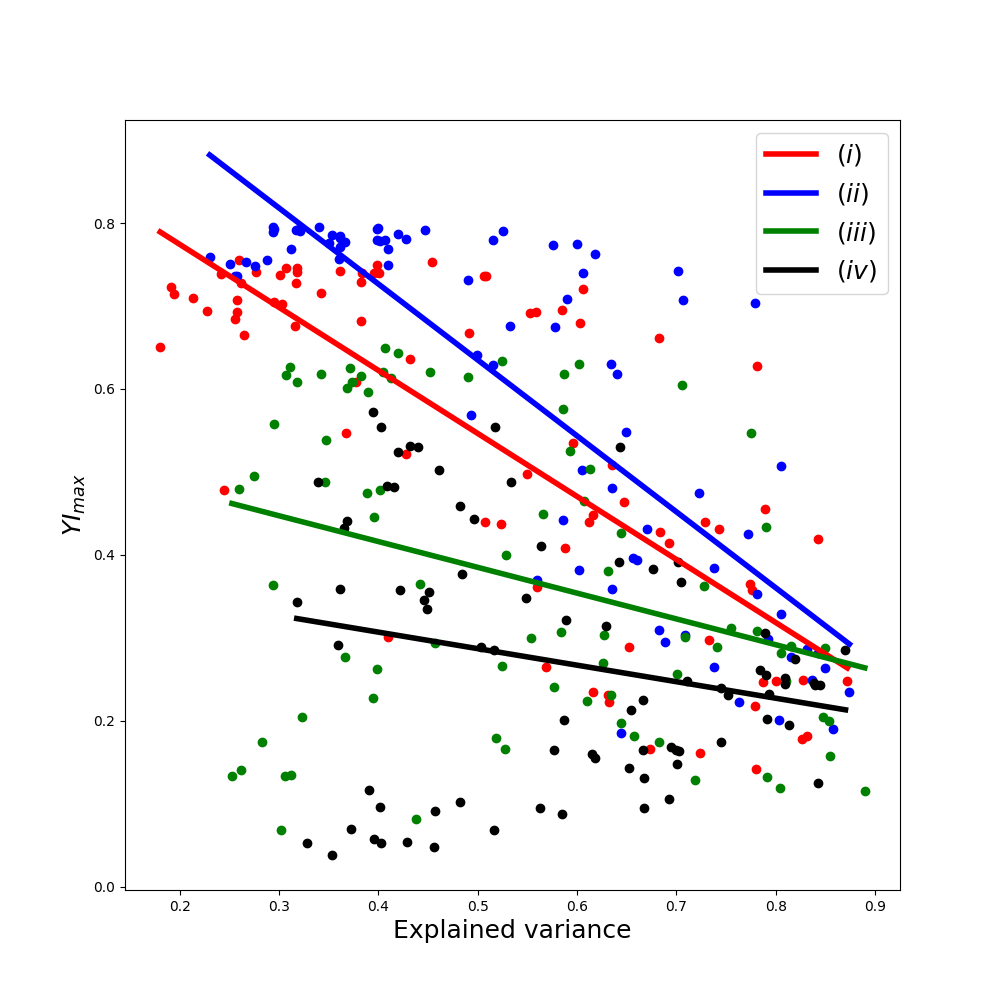}
         \caption{Complex PT models.}
         \label{fig:EVvsYImax_complex}
     \end{subfigure}
        \caption{Relation between explained variance using 10 principal components and model performance represented as $YI_{max}$.}
        \label{fig:EVvsYImax}
\end{figure*}

Figure \ref{fig:EVvsSensitivity} represents the explained variance against sensitivity. We can see that the trend is flat for strategy \textit{(iv)}, but positive for strategies \textit{(i)-(iii)}. This means that the trends in Figure \ref{fig:EVvsYI} are explained by specificity rather than sensitivity.
By balancing sensitivity and specificity, \ie $YI_{max}$ as seen in Figure \ref{fig:EVvsYImax}, the rate of change is reduced compared to $YI$ in Figure \ref{fig:EVvsYI}. 


\subsection{Example predictions}
Table \ref{tab:example_predictions} 
shows a few examples of correct (TP and TN) and incorrect predictions (FN and FP).

\begin{table*}[t]
    \centering
    \begin{tabular}{|l|l|r|r|r|c|}\hline 
        \textbf{Chemical} & \textbf{Species} & $\mathbf{\log(\kappa)}$ & \textbf{Predicted} & \textbf{Lethal} & \textbf{Classification} \\\hline\hline
        
        D001556 (hexachlorocyclohexane) & 59899 (walking catfish) & $-3.4$ & $0.97$ & $1$ (yes) & TP \\
        C037925 (benthiocarb) & 7965 (sea urchins) & $0.9$ & $0.2$ & $0$ (no) & TN \\
        D026023 (permethrin) & 378420 (bivalves) & $0.7$ & $0.96$ & $1$ (yes) & TP \\\hline\hline 
        D011189 (potassium chloride) & 938113 (megacyclops viridis) & $6.7$ & $0.27$ & $1$ (yes) & FN\\
        C427526 (carfentrazone-ethyl) & 208866 (eudicots) & $-0.9$ & $0.82$ & $0$ (no) & FP\\
        D010278 (parathion) & 201691 (green sunfish) & $-0.9$ & $0.86$ & $0$ (no) & FP\\\hline
    \end{tabular}
    \vspace{0.2cm}
    \caption{Example predictions by Complex FT HolE-DistMult (best model) for sampling strategy \textit{(iv)}.}
    \label{tab:example_predictions}
\end{table*}

\textit{Benthiocarb} and \textit{permethrin} are both biocides with different targets: \textit{benthiocarb} is a herbicide and \textit{permethrin} is an insecticide. It is therefore not surprising that \textit{benthiocarb} has a low predicted effect on sea urchins, while \textit{permethrin} has a severe effect on bivalves. 

There are several possible explanations for the failed predictions.
A wrong prediction of \textit{potassium chloride} toxicity to a marine copepod (\textit{Megacyclops viridis}) could be due to the prediction model not being accurate enough for metal salts, or the copepod species being particularly sensitive to changes in osmolarity due to salt content. The wrong prediction of lack of herbicide toxicity (\ie \textit{carfentrazone-ethyl}) to a flower (\ie \textit{eudicots})  could be due to the fact that flowers, and plants in general, are severely underrepresented in the available effect prediction data.

\section{Discussion}
\label{sec:discussion}

We have introduced the Toxicological Effect and Risk Assessment (TERA) knowledge graph and shown how we can directly use it in chemical effect prediction. The use of TERA 
improves the PT-based prediction models over the one-hot 
baselines. In the most challenging data sampling strategies, we have also 
seen the benefits of 
creating tailored (\ie fine-tuned)
KG embeddings in the FT-based prediction models. 

\subsection{TERA knowledge graph}
\label{sec:disc_tera}

The constructed knowledge graph 
consists of several sources from the ecotoxicological domain. There are three major parts 
in TERA: the effects data, the chemical data, and the species taxonomic data. Integrating each part 
has different challenges. The chemical and pharmacological communities 
have come a long way in annotating their data as knowledge graphs and ontologies. Here, selecting the correct subsets to work with the chemical effect prediction data was a major challenge. This had to be done based on mappings between effect data and chemical data that were extracted from  Wikidata.
We selected a 
relatively small subset of the chemical sub-KG to facilitate faster model training, however, still 
larger than the extracted fragment from the species sub-KG. The species sub-KG was created from tabular data and cleaned by removing several annotation labels with redundant information. This sub-KG was aligned using ontology alignment systems to the species taxonomy in the effects sub-KG. This required pre-processing of the KG, where it was divided into smaller parts such that the selected systems could perform the alignment. We used several standard ontologies to facilitate the transformation of the effect data into a knowledge graph. This involved not only automatic processes, but also an important amount of manual work.
 


\change{
Integrating more data into TERA involves the creation of mappings to the existing data. This is possible for a large amount of chemical datasets as Wikidata links multiple datasets, \eg the chemical compound diethyltoluamide (\texttt{wd:Q408389}) has $\sim 35$ distinct identifiers. 
Biological data, both taxonomic and effects, might be harder to align to TERA as these mappings are not available in Wikidata. Here, ontology alignment systems play an important role to fill this gap.

The additional integrated data will give larger coverage of the domain, and thereby, improve model performance. However, adding more data will also increase the memory and time requirements of KGE models. This was bypassed in this work by reducing TERA to only relevant parts.

Adding additional domain knowledge is also critical in other applications, such as using TERA for data access. 
}

\subsection{Performance of prediction models}
We have 
shown that 
the ability to embed some structure types
of different KGE models largely impact the prediction models. We see that some KGE models fail to capture the semantics of the chemicals and the species, which leads to similar performance to the one-hot baselines. Moreover, in a few isolated cases the performance is reduced further which leads us to believe that the embeddings \emph{collapse} in one or 
some dimensions, making it impossible to distinguish among entities. 

We suspect that the even distribution of KGE models to embed $KG_C$ (Table \ref{tab:best_models}) in most settings is likely down to the structure of $KG_C$. This sub-KG has, unlike $KG_S$'s tree structure, a forest structure, and models that can deal with trees (as in $KG_S$) fail here, \eg 
an entity in $KG_C$ can have multiple parents, but only one grand-parent. 
In this case, some models may
create very similar or the same embeddings for the parent nodes.

\section{Conclusions and future work}
\label{sec:conclusions}

TERA is a novel knowledge graph which includes large amounts of data required by ecological risk assessment.
We have conducted an extensive evaluation of KGE embedding models in a novel and very challenging application domain. Moreover, we have shown the value of using TERA in an ecotoxicological effect prediction task.
The fine-tuning optimization model architecture to adapt the KG embeddings to the prediction task has, to our knowledge, not been applied elsewhere.


\subsection{Value for the ecotoxicology community}
\change{
The creation of TERA is of great importance to future effect modelling and computational risk assessment approaches within ecotoxicology. Where the strategic goal is designing and developing prediction models to assess the hazard and risks of chemicals and their mixtures where traditional laboratory data cannot easily be acquired.

A great effort in the hazard and risk assessment of chemicals is the reduction of regulatory-mandated animal testing. Wide-scale predictive approaches, as described here, answer a direct and current need for generalized prediction frameworks. These can aid in identifying especially sensitive species and toxic chemicals. At the Norwegian Institute for Water Research (NIVA), TERA will be used in this regard and will support several research projects. 

In environmental risk assessment it is often unfeasible to assess the hazard and risk a chemical poses to a local species in the environment. These species may not be suitable for lab testing, or may even be endangered and thus are protected by national or international legislation. The currently presented work provides an in silico approach to predict the hazard to such species based on the taxonomic position of the species within the tree of life.

From an economic perspective, TERA and the prediction models are useful tools to evaluate new industrial chemicals during the synthetic in silico stage. Candidate chemicals can be evaluated for their potential environmental hazard, which is in line with the Green Chemistry initiatives by authorities such as the European Parliament or the US Environmental Protection Agency. 
}

The effect prediction using TERA is also in line with a larger shift in ecological risk assessment towards the use of artificial intelligence~\cite{WITTWEHR2019100114}.
We also believe the development of TERA contributes to a methodological change in the community, and encourages others to make their data interoperable. 


\subsection{TERA as background knowledge}
As mentioned, in this work we use TERA directly in prediction models. However, TERA could be used as background knowledge to improve many emerging techniques for toxicity prediction (\eg \cite{Sharma2017}). These methods often use chemical features, images, fingerprints and so on as input, and machine learning methods such as Convolutional Neural Networks and Random Forests as prediction models \cite{Wu2018,yang2018silico}. These models are often uninterpretable, and the predictions lack domain explanations. 
TERA can also provide context
for machine learning tasks such as pre-processing, feature extraction, transfer and zero/few-shot learning. Furthermore, the knowledge graph is a possible source for the (semantic) explanation of the predictions (\eg~\cite{DBLP:journals/corr/abs-1805-10587}).

\subsection{Benchmarking KG embedding models}

We have shown that embedding TERA brings new challenges to state-of-the-art KGE models with respect to capturing the semantics of the chemicals and the species. Furthermore, as shown in Section \ref{sec:teraprediction} the sparsity-related measures indicate that TERA represent an interesting KG.  KGE models could be benchmarked in a standard KG completion task or in a specific task such as the chemical effect prediction.

\subsection{Value to the ontology alignment community}
As mentioned in Section \ref{sec:datapreparation}, there does not exist a complete and public alignment between ECOTOX species and the NCBI Taxonomy. 
Therefore the computed mappings can also be seen as a very relevant resource to the ecotoxicology community. 
The used alignment techniques achieve high scores for recall over the available (incomplete) reference mappings. 
However, aligning such large and challenging datasets requires preprocessing before ontology alignment systems can cope with them.
We removed all nodes which did not share a word (or shared only a stop word) in labels across the two taxonomies. 
This quartered the size of ECOTOX and reduced NCBI Taxonomy 50 fold. However, the possible alignment between entities without labels is lost when reducing the dataset size. 
Thus, the alignment of ECOTOX and NCBI Taxonomy has the potential of becoming a new track of the Ontology Alignment Evaluation Initiative (OAEI)~\cite{oa_ecotox2020} to push the limits of large scale ontology alignment tools. Furthermore, the output of the different OAEI participants could be merged into a rich consensus alignment (\eg as done in the phenotype-disease domain \cite{DBLP:journals/biomedsem/HarrowJSRWMAKMW17}) that could become the reference alignment to integrate ECOTOX~and~NCBI~Taxonomy. 


\subsection{Future work}
We plan to extend TERA to include a larger part of ChEBI (which ChEMBL is a part of). ChEBI includes relevant data on the interaction between chemicals and species at a cellular level, which may be very important for chemical effect prediction. In this work we only consider effect data from ECOTOX as this is the largest data set available, however, the inclusion of \eg TOXCAST \cite{toxcast} is in our interest. \change{New sources will always bring more coverage of the domain and will improve TERA for prediction, as background knowledge, and for data access.}

We plan to evaluate the effect prediction under different parts of TERA, \ie which sources in TERA provide value and which do not contribute in terms of the effect prediction. \change{A similar effort in exploring different KG crawling techniques has been explored in \cite{npmasterthesis}. In a similar vain, we plan to evaluate how materialization, via OWL reasoning, of TERA's implicit triples affects prediction performance. 
}

Finally, as mentioned already, some KGE models cannot deal with parts of the structure of TERA. An in-depth analysis of this is an interesting direction for future research. This could be solved by embedding the hierarchy separately, \eg 
\cite{Mumtaz2021},
\change{or imposing restrictions on the embeddings, such as a minimum distance constraint.}


\subsection{Resources}
We encourage feedback from domain researchers on extensions to TERA and associated tools. 

\smallskip
\noindent
A snapshot of TERA is available at 
\begin{center}
\url{https://doi.org/10.5281/zenodo.3559865}
\end{center}
This snapshot does not include data that is impractical to re-share (\ie partial $KG_C$ as described in Section \ref{sec:tera}). However, we include the full $KG_E$ and $KG_S$. 

\smallskip
\noindent
All the material related to this project is available at
\begin{center}
\url{https://github.com/NIVA-Knowledge-Graph/}
\end{center}

Source codes to create TERA are available in the \href{https://github.com/NIVA-Knowledge-Graph/TERA}{\emph{TERA}} GitHub repository.
The prediction models and data used for prediction can be found in the 
\href{https://github.com/NIVA-Knowledge-Graph/KGs_and_Effect_Prediction_2020}{\emph{KGs\_and\_Effect\_Prediction\_2020}} GitHub repository.
The prediction models require the implementation of the KGE models from the \href{https://github.com/NIVA-Knowledge-Graph/KGE-Keras}{\emph{KGE-Keras}} GitHub repository.

\section*{Acknowledgements}

This work is supported by the grant 272414 from the Research Council of Norway (RCN), the MixRisk project (Research Council of Norway, project 268294), SIRIUS Centre for Scalable Data Access (Research Council of Norway, project 237889), Samsung Research UK, Siemens AG, and the EPSRC projects AnaLOG (EP/P025943/1), OASIS (EP/S032347/1), UK FIRES (EP/S019111/1) and the AIDA project (Alan Turing Institute).

\bibliographystyle{abbrv}
\bibliography{bibliography}

\appendix
\section{Knowledge Graph Embedding Models}
\label{sec:appendix_kge}



In this work, we use 9 KGE models of three major categories: decomposition models, geometric models, and convolutional models. The interested reader please refer to \cite{DBLP:journals/tkdd/RossiBFMM21} for a comprehensive survey.

\subsection{Notation}
Throughout this section we use bold letters to denote vectors while matrices are denoted as $\mat{M}$. Common notation for all KGE models are, $\norm{\cdot}_n$ for the $n$-th norm, $\inner{\vec{x},\vec{y}}$ for the inner product (dot product) between $\vec{x}$ and $\vec{y}$, $[\vec{x};\vec{y}]$ is the concatenation of $\vec{x}$ and $\vec{y}$, $\overline{\vec{x}}$ indicates the reshape of a one-dimensional vector into a two-dimensional \textit{image} (\emph{not} in HolE where it represent the complex conjugate), finally, $\func{vec}(\mat{X})$ reshapes a matrix into a one-dimensional vector.

The vector representation of an entity and a relation are noted as $\vec{e}_e$ and $\vec{e}_p$, respectively. These vectors are either in $\mathbb{R}^k$ or $\mathbb{C}^k$, where $k$ is the embedding dimension.

\subsection{Decomposition models}
\noindent
\textbf{DistMult.}
Developed by \cite{Yang2015EmbeddingEA} and shown to have state-of-the-art performance on link prediction tasks under optimal hyper-parameters \cite{DBLP:journals/corr/KadlecBK17}. This model represent the score of a triple as an Hadaman product (dot product) of the vectors representing the subject, predicate, and object of a triple. 
\begin{align}
    SF_{\text{DistMult}}(sb,p,ob) = \inner{\vec{e}_{sb},\vec{e}_p,\vec{e}_{ob}}
\end{align}
This model does not take the direction of the relation into account, that is, $SF_{\text{DistMult}}(sb,p,ob)=SF_{\text{DistMult}}(ob,p,sb)$.

\medskip
\noindent
\textbf{ComplEx.}
This model use the same scoring function as DistMult \cite{DBLP:journals/corr/TrouillonWRGB16}. However, the entity vector representation are in the complex space ($\vec{e}_{sb}, \vec{e}_p, \vec{e}_{ob}\in \mathbb{C}^k$) and therefore, the drawback of lacking directionality in DistMult is solved. 
\begin{align}
    \begin{split}
    SF_{\text{ComplEx}}(sb,p,ob) &= \inner{\vec{e}_{sb},\vec{e}_p,\vec{e}_{ob}}\\ 
    &= \inner{\Re(\vec{e}_{sb})+i\Im(\vec{e}_{sb}),\Re(\vec{e}_{sb})}\\
    &+ \inner{i\Im(\vec{e}_{sb}),\Re(\vec{e}_p)+i\Im(\vec{e}_{ob})} \\
    &= \inner{\Re(\vec{e}_{sb}),\Re(\vec{e}_p),\Re(\vec{e}_{ob})} \\
    &+ \inner{\Im(\vec{e}_{sb}),\Re(\vec{e}_p),\Im(\vec{e}_{ob})} \\
    &+ \inner{\Re(\vec{e}_{sb}),\Im(\vec{e}_p),\Im(\vec{e}_{ob})} \\
    &+ \inner{\Im(\vec{e}_{sb}),\Im(\vec{e}_p),\Re(\vec{e}_{ob})} 
    \end{split}
\end{align}
where $i=\sqrt{-1}$ and, $\Re(x)$ and $\Im(x)$ are the real and complex parts of $x$, respectively. We can easily see that $SF_{\text{ComplEx}}(\vec{e}_{sb},\vec{e}_p,\vec{e}_{ob})=SF_{\text{DistMult}}(\vec{e}_{sb},\vec{e}_p,\vec{e}_{ob})$ if $\Im(\vec{e}_{sb})=\Im(\vec{e}_p)=\Im(\vec{e}_{ob})=~\vec{0}$.

\medskip
\noindent
\textbf{HolE.}
The Holographic embedding model is described in \cite{DBLP:journals/corr/NickelRP15}, and use a circular correlation scoring function
\begin{align}
    SF_{\text{HolE}}(sb,p,ob) &= \vec{e}_p^T [\vec{e}_{sb} \star \vec{e}_{ob}], \\
    \vec{e}_{sb} \star \vec{e}_{ob} &= \mathcal{F}^{-1}[\overline{\mathcal{F}(\vec{e}_{sb})}\circ \mathcal{F}(\vec{e}_{ob})]
\end{align}
where $\mathcal{F}$ and $\mathcal{F}^{-1}$ are the Fourier transform and its inverse, for this model we use $\overline{x}$ as the elementwise complex conjugate, $\circ$ denotes the Hadamard product (element-wise). HolE has been show to be equivalent to ComplEx \cite{hayashi-shimbo-2017-equivalence}, and therefore, we expect the performance to be similar. 

\subsection{Geometric models}
\noindent
\textbf{TransE.}
The translational model has the scoring function \cite{NIPS2013_5071}
\begin{align}
    SF_{\text{TransE}}(sb,p,ob) = ||\vec{e}_{sb} + \vec{e}_p - \vec{e}_{ob}||_n .
\end{align}
Such that if $(sb,p,ob)$ exists in the KG the relational embedding will translate the subject embedding close to the object embedding. 

\medskip
\noindent
\textbf{RotatE.} This model is inspired by Euler's identity ($e^{i\theta}=\cos(\theta)+i\sin(\theta)$) and scores triples by rotating the relation embedding in complex space. RotatE has been shown to be efficient of modelling symmetric, inverse and composite relations \cite{sun2018rotate}.
The scoring function of RotatE is defined as
\begin{align}
    \begin{split}
    &SF_{\text{RotatE}}(sb,p,ob) = ||\vec{e}_{sb} \circ \vec{e}_p - \vec{e}_{ob}||_n \\
    &= ||\vec{e}_{sb} \circ (\cos(\vec{\theta}_p)+i\sin(\vec{\theta}_p))-\vec{e}_{ob}||_p\\ 
    &= ||[\Re(\vec{e}_{sb})\cos(\vec{\theta}_p)-\Im(\vec{e_{sb}})\sin(\vec{\theta}_p)-\Re(\vec{e}_{ob})\\
    &\quad ;\Re(\vec{e}_{sb})\sin(\vec{\theta}_p)+\Im(\vec{e_{sb}})\cos(\vec{\theta}_p)-\Im(\vec{e}_{ob})]||_n .
    \end{split}
\end{align}
Here, we concatenate the real and complex parts of $\vec{e}_{sb} \circ \vec{e}_p - \vec{e}_{ob}$. The modulus constraint of $\vec{e}_p$ is set equal to $1$ and is therefore not included in the scoring function. 
See the original publication for details of derivation. 

\medskip
\noindent
\textbf{pRotatE.}
This model is described as a baseline for RotatE enabling comparison when including modulus information in the model versus limiting to phase information only \cite{sun2018rotate}. pRotatE has the scoring function
\begin{align}
    SF_{\text{RotatE}}(sb,p,ob) &= 2MC||\sin(\frac{\vec{\theta}_{sb} + \vec{\theta}_p - \vec{\theta}_{ob}}{2})||_n
\end{align}
where $\vec{\theta}_x = \angle \vec{e}_x$ (phase of $\vec{e}_x$) and $MC$ is the modulus constraint on $\vec{e}_{sb}$ and $\vec{e}_{ob}$.

\medskip
\noindent
\textbf{HAKE.} The hierarchy-aware model use the modulus and the phase part of the embedding vectors \cite{zhang2019learning}. Such that entities at the same level in the hierarchy is modelled using rotation, \ie phase, and the entities at different levels are modelled using the distance from the origin, \ie modulus. Therefore, the scoring function of HAKE is modelled in two parts
\begin{align}
    SF_{\text{pRotatE}}(sb,p,ob) &= |||\vec{e}_{sb}|\circ |\vec{e}_p|-|\vec{e}_{ob}|||_n \nonumber \\&+ ||\sin(\frac{\vec{\theta}_{sb} + \vec{\theta}_p - \vec{\theta}_{ob}}{2})||_1
\end{align}
where $|\cdot|$ is the modulus of $\cdot$. The authors noted that a mixture bias can be added to $|||\vec{e}_{sb}|\circ |\vec{e}_p|-|\vec{e}_{ob}|||_n$ to improved performance \cite{zhang2019learning}. We omit these details here.

\subsection{Convolutional models}
The final set of models used in this work are convolutional models. We denote convolutions between an \textit{image} $X$ and filters $\omega$ is denoted as $X\ast \omega$. The models also use dense layers, which is denoted by transform matrices, \eg $\mat{W}$, note that this also includes bias, even though we do not explicit state it. Moreover, dropout layers are used between every convolutional and dense layer.

\medskip
\noindent
\textbf{ConvKB.}
The scoring function of ConvKB \cite{Nguyen2018} use a single convolutional layer and a single dense layer
\begin{align}
    SF_{\text{ConvKB}}(sb,p,ob) & = & \nonumber \\ & f(\func{vec}(f([\vec{e}_{sb};\vec{e}_p;\vec{e}_{ob}]\ast \omega))\mat{W}),
\end{align}
where $\func{vec}(x)$ reshapes $x$ to a 1-dimensional vector. $\omega$ is the convolution filters.  
$\mat{W}$ is the transformation matrix for the output dense layer. ConvKB can easily be extended to use multiple convolution and dense layers. 

\medskip
\noindent
\textbf{ConvE.}
In contrast to ConvKB, ConvE \cite{dettmers2018} only perform convolution over the subject and predicate \textit{image} (concatenated and reshaped) and multiples the output dense layer with the object vector as such
\begin{align}
    SF_{\text{ConvE}}(sb,p,ob) & = & \nonumber \\ & f(\func{vec}(f([\overline{\vec{e}_{sb}};\overline{\vec{e}_p}]\ast \omega))\mat{W})\vec{e}_{ob}^T
\end{align}
where $\overline{\vec{x}}$ reshapes $\vec{x}$ into a 2-dimensional \textit{image}. 
Here, the last dimension of $\mat{W}$ is equal to the embedding dimension. This model can also be extended with multiple convolution and dense layers, however, \cite{dettmers2018} found that this did not yield improved results. 

\subsection{Loss functions}
\label{sec:loss_functions}
Work on KGE models usually define loss functions specific to the models. However, as show in \cite{nayyeri2019understanding,MohamedetalDLKG} the choice of loss function has a huge impact on model performance. In this work we use four loss functions. We experimented with other loss functions, \eg absolute/square error, however, these did not materialize in improved results.

To optimize a loss function we need to generate negative examples. Under the local closed world assumption we replace the object of each true triple with all entities and sample negative examples from this set~\cite{45634}, \ie we sample from $\{sb,p,ob^\prime\} \not\in KG, ob^\prime \in \mathcal{E}$. This can be expanded to the stochastic local closed world assumption, which corrupt both the subject and the object of true triples (illustrated by Fig. 3 in \cite{ali2020bringing}). The number of negative samples sampled per positive sample is controlled by a hyper-parameter. However, \cite{DBLP:journals/corr/KadlecBK17} show that the largest possible number is favorable. 

\medskip
\noindent
\textbf{Pointwize hinge.}
The objective of pointwize losses minimize the scores of negative triples and maximize the score of positive triples. 
\begin{align}
    L_{H_1} = \sum_{\vec{t}\in X} [\gamma - y_{\vec{t}} S(\vec{t})]_+
\end{align}
where $X$ is the set of positive and negative triples, $y$ is the triple label ($-1$ for false and $1$ for true) and $S(\vec{t})$ is the score of triple $\vec{t}$. $\gamma$ is the margin hyper-parameter. $[x]_+$ is the positive part of $x$.

\medskip
\noindent
\textbf{Pointwize logistic.}
In contrast to hinge loss, logistic loss applies a larger non-linear loss to predictions that are further away from the true label. 
\begin{align}
    L_{L_1} = \sum_{t\in X} \func{log}(1+\func{exp}(-y_{\vec{t}} S(\vec{t}))
\end{align}

\medskip
\noindent
\textbf{Pairwise hinge.}
The objective of pairwise loss functions is to maximize the distance (in score) between a positive and a negative triple. 
\begin{align}
    L_{H_2} = \sum_{\vec{t}^+\in X^+}\sum_{\vec{t}^-\in X^-} [\gamma + S(\vec{x}^-) - S(\vec{x}^+)]_+
\end{align}
where $X^+$ and $X^-$ are the sets of positive and negative triples, respectively. $\gamma$ is the margin hyper-parameter, which for pairwise hinge loss represents the maximum score discrepancy between a positive and negative score. 

\medskip
\noindent
\textbf{Pairwise logistic.}
Akin to the move from pointwize to pairwize hinge, pairwize logistic maximizes the distance between positive and negative triples, however, in a non-linear way 
\begin{align}
    L_{L_2} = \sum_{\vec{t}^+\in X^+}\sum_{\vec{t}^-\in X^-} \func{log}(1+\func{exp}(S(\vec{x}^-) - S(\vec{x}^+)) .
\end{align}

\subsection{Implementation}
We have implemented the KGE models in Keras~\cite{chollet2015keras} and the model codes are available at \url{https://github.com/NIVA-Knowledge-Graph/KGE-Keras}. This enables us to easily use the KGE models as components in other models as described in Section \ref{sec:prediction}.

\end{document}